\documentclass[lettersize,journal]{IEEEtran}
\usepackage{amsmath,amsfonts}
\usepackage{algorithm}
\usepackage{algorithmicx}
\usepackage{algpseudocode}
\makeatletter
\def\algbackskip{\hskip-\ALG@thistlm}
\makeatother

\usepackage{array}
\usepackage[caption=false,font=normalsize,labelfont=sf,textfont=sf]{subfig}
\usepackage{textcomp}
\usepackage{stfloats}
\usepackage{url}
\usepackage{verbatim}
\usepackage{graphicx}
\usepackage{multirow}
\usepackage{xcolor}
\usepackage{amssymb}
\usepackage{enumitem}
\usepackage[flushleft]{threeparttable}
\usepackage{caption}

\usepackage{mwe}
\usepackage{float}
\def\BibTeX{{\rm B\kern-.05em{\sc i\kern-.025em b}\kern-.08em
    T\kern-.1667em\lower.7ex\hbox{E}\kern-.125emX}}
\usepackage{balance}
\newtheorem{theorem}{Theorem}

\newtheorem{definition}{Definition}
\newcommand{\pname}{{Hat-DFed}\xspace}
\begin{document}
\title{Towards Heterogeneity-Aware and Energy-Efficient Topology Optimization for Decentralized Federated Learning in Edge Environment}

\author{Yuze~Liu,
        Tiehua~Zhang,~\IEEEmembership{Member,~IEEE,}
        Zhishu~Shen,~\IEEEmembership{Member,~IEEE,}
        Libing~Wu,~\IEEEmembership{Senior Member,~IEEE,}
        Shiping~Chen,~\IEEEmembership{Senior Member,~IEEE,} and
	    Jiong~Jin,~\IEEEmembership{Member,~IEEE}

\thanks{\textit{Corresponding author: Tiehua Zhang.}}
\thanks{Yuze Liu and Jiong Jin are with the School of Engineering, Swinburne University of Technology, Melbourne, Australia (e-mail: \{yuzeliu, jiongjin\}@swin.edu.au).}
\thanks{Tiehua Zhang is with the School of Computer Science and Technology, Tongji University, Shanghai, China, and also with State Key Lab for Novel Software Technology, Nanjing University, Nanjing, China (e-mail: tiehuaz@tongji.edu.cn).}
\thanks{Zhishu Shen is with the School of Computer Science and Artificial Intelligence, Wuhan University of Technology, Wuhan, China (e-mail: z\_shen@ieee.org).}
\thanks{Libing Wu is with the School of Cyber Science and Engineering, Wuhan University, Wuhan, China (e-mail: wu@whu.edu.cn).}
\thanks{Shiping Chen is with the CSIRO Data61, Sydney, Australia (e-mail: shiping.chen@data61.csiro.au)}

}


\maketitle

\begin{abstract}
Federated learning (FL) has emerged as a promising paradigm within edge computing (EC) systems, enabling numerous edge devices to collaboratively train artificial intelligence (AI) models while maintaining data privacy. To overcome the communication bottlenecks associated with centralized parameter servers, decentralized federated learning (DFL), which leverages peer-to-peer (P2P) communication, has been extensively explored in the research community. Although researchers design a variety of DFL approach to ensure model convergence, its iterative learning process inevitably incurs considerable cost along with the growth of model complexity and the number of participants. These costs are largely influenced by the dynamic changes of topology in each training round, particularly its sparsity and connectivity conditions. Furthermore, the inherent resources heterogeneity in the edge environments affects energy efficiency of learning process, while data heterogeneity
degrades model performance. These factors pose significant challenges to the design of an effective DFL framework for EC systems.
To this end, we propose Hat-DFed, a heterogeneity-aware and coset-effective decentralized federated learning (DFL) framework. In \pname, the topology construction is formulated as a dual optimization problem, which is then proven to be NP-hard, with the goal of maximizing model performance while minimizing cumulative energy consumption in complex edge environments.
To solve this problem, we design a two-phase algorithm that dynamically constructs optimal communication topologies while unbiasedly estimating their impact on both model performance and energy cost. Additionally, the algorithm incorporates an importance-aware model aggregation mechanism to mitigate performance degradation caused by data heterogeneity. Extensive experiments demonstrate that \pname outperforms state-of-the-art baselines, achieving an average 1.9\% improvement in test accuracy while reducing total energy cost by 36.7\% throughout learning process.

\end{abstract}


\section{Introduction}
\IEEEPARstart{O}{ver} the last decade, the advancement of Internet of Things (IoT) technologies has resulted in a surge of edge devices, such as smart sensors and mobile phones, generating colossal amount of data from various applications~\cite{stoyanova2020survey}. Meanwhile, deep learning (DL) models have become fundamental to a wide range of applications deployed on edge devices, such as face recognition~\cite{luo2021binarized} and traffic signal control~\cite{wang2025towards}, which typically require substantial amounts of training data. Edge computing (EC) has emerged as a promising paradigm to bridge the gap between the IoT and AI by bringing computational capability closer to data sources, thereby significantly reducing transmission costs compared to traditional cloud-based architectures~\cite{li2019edge}. However, how to leverage the rich data generated in edge environments to enhance DL model performance, while ensuring data privacy, remains a critical challenge in EC systems~\cite{nadella2024advancing}.


As a privacy-aware distributed learning paradigm, federated learning (FL) offers an effective solution by enabling collaborative model training across participants without data sharing. Prior research~\cite{mcmahan2017communication,zhang2023adaptive,wang2025tdml} investigates centralized FL (CFL) architectures, wherein a central parameter server (PS) coordinates the learning process by collecting client updates, performing aggregation, and distributing the updated global model. However, the PS in CFL architectures incurs substantially higher computational and communication overhead compared to edge participants, making it often impractical for resource-constrained EC environments~\cite{chen2022decentralized}. Decentralized federated learning (DFL), which leverages peer-to-peer (P2P) communication, is a promising architecture for enabling collaborative learning without the need for resource-intensive central servers in EC systems. In such protocol and architecture, each participant trains locally and aggregates models only from neighboring nodes orchestrated by designated algorithms, ensuring bounded communication costs and making DFL highly scalable for resource-constrained EC environments.

Previous studies~\cite{wu2024topology,neglia2020decentralized} have shown that developing efficient DFL methods necessitates consideration of two core processes: dynamic construction of communication topology and robust model parameter aggregations. For instance, DFL methods addressing communication topology construction~\cite{DPSGD,lee2020tornadoaggregate} investigate how various network structures influence model performance and communication overhead during collaborative learning, particularly as the complexity of models and the number of participants increase. Conversely, methods that focus on the model aggregation process~\cite{pushsum,zantedeschi2020fully} emphasize designing updating mechanisms to improve model accuracy. Both the communication topology and aggregation strategy significantly impact the final model performance and energy consumption over the course of training. These two design aspects are closely aligned with the fundamental goals of deploying DFL in EC systems.

Despite its potential, designing an efficient DFL for practical EC systems still faces two key challenges:\begin{itemize}
    \item \textit{System Heterogeneity.} Each participant in EC systems exhibits diverse computational and communication capabilities due to varying hardware specifications~\cite{yang2021characterizing}. This disparity leads to imbalanced energy consumption, with less energy-efficient devices expending significantly more energy than more efficient devices when executing identical tasks~\cite{zeng2022gnn}.
    \item \textit{Data Heterogeneity.} Training data reflects local environmental conditions and device usage patterns, resulting in non-IID distributions across participants~\cite{xu2024overcoming}. In addition, unpredictable device churn like intermittent connectivity~\cite{guo2021towards} causes time-varying data heterogeneity, causing dynamic variations in both data volume and data distributions throughout the learning process. While both forms of data heterogeneity degrade model performance, temporal variations additionally affect energy cost during the learning process.
\end{itemize} 

For the recent studies addressing topology optimization in heterogeneous edge environments, they exhibit divergent limitations. Zeng \textit{et al.}~\cite{zeng2022gnn} and Wang \textit{et al.}~\cite{wang2020service} minimized energy costs through topology construction but neglect model performance. Zec \textit{et al.}~\cite{DAC} and Onoszko \textit{et al.}~\cite{onoszko2021decentralized} optimized neighbor selection for non-IID data to enhance accuracy, yet disregard energy efficiency. It is essential to design an efficient DFL framework for practical EC deployments by simultaneously addressing these dual challenges, with the goal of maximizing model performance while minimizing energy consumption.
To this end, we propose a \underline{H}eterogeneity-\underline{a}ware and energy-efficien\underline{t} DFL framework, called \pname, to enable efficient collaborative learning in practical EC systems. In \pname, We formulate a joint optimization problem that explicitly analyzes the dual impact of communication topology on model performance and energy cost in heterogeneous edge environments. This impact is quantified using a novel utility-based evaluation metric, and the problem is proven to be NP-hard. To solve this NP-hard optimization problem, we design a two-phase algorithm comprising Utility-based Topology Construction (UTC) and Decentralized Collaborative Model Update (DCMU). Specifically, UTC estimates the utility of communication links based on historical profiles of model performance and energy cost, and generates optimized topologies via a utility-guided link selection algorithm. Once the communication topology is established, DCMU is executed to enable collaborative model updating through a sequence of local training, model exchange, and importance-aware model aggregation.



In summary, the main contributions of this work are:
\begin{itemize}
    \item We propose \pname, a novel DFL framework designed to enable efficient collaborative learning in heterogeneous EC systems. In \pname, topology construction is formulated as a joint optimization problem that aims to maximize model accuracy while minimizing cumulative energy consumption. To capture the trade-off between both objectives, we introduce an evaluation metric, termed \textit{utility}, to quantify the dual impact of each topology connection on model performance and energy cost. We further prove that this problem is NP-hard.

    \item We design an effective two-phase algorithm to solve the formulated optimization problem. By constructing an optimal communication topology based on estimated utility, our algorithm effectively balances model performance and energy efficiency. Furthermore, we introduce an importance-aware model aggregation method to counteract performance degradation caused by non-IID and time-varying data heterogeneity across edge participants during learning process.
    \item We conduct extensive experiments on various tasks using the Fashion-MNIST and CIFAR-10 datasets to evaluate both model performance and energy efficiency. The simulation results demonstrate that our proposed framework significantly outperforms state-of-the-art baselines in terms of both accuracy and energy efficiency throughout the collaborative learning process.
    \item We have released our source code publicly at https://github.com/papercode-DFL/Hat-DFed to facilitate further research and development in this field.
\end{itemize}
The remainder of this paper is organized as follows. Section~\ref{sec:2} reviews related work on CFL and DFL in EC systems. Section~\ref{sec:3} introduces the preliminaries of our EC system and DFL. Section~\ref{sec:4} formulates the topology construction problem and proves its NP-hardness, while Section~\ref{sec:5} presents the proposed two-phase algorithm in detail. Experimental results are reported and analyzed in Section~\ref{sec:6}. Finally, Section~\ref{sec:7} concludes the paper, and the appendix provides the proofs of Theorem 1 and 2.


\section{Related Work}
\label{sec:2}


With the advent of EC, interconnected systems and devices in edge environments can collaboratively perform data collection, processing, and analysis close to the data source. Meanwhile, recent studies have raised increasing concerns about ensuring data privacy during the collaborative learning process.~\cite{xu2021edge,wu2024fedcache}. To address this concern, FL has been introduced as a timely solution for EC systems, aiming to protect participants’ local data privacy while achieving superior model performance. A research direction~\cite{mcmahan2017communication,ji2019learning,zhang2023adaptive} focuses on adopting CFL architecture with a central PS. One of the widely used CFL methods is FedAvg~\cite{mcmahan2017communication}, in which the local model weights are sent to the PS and aggregated in an iterative manner. FedAtt~\cite{ji2019learning} assigns weighting coefficients to each local model during aggregation, based on the distance between participant models and the global model at each communication round. However, with increasing model complexity and a growing number of participants, resource bottlenecks can occur at the PS, resulting in poor scalability, especially in resource-limited EC systems~\cite{chen2022decentralized}.

To address the scalability issue, DFL has attracted growing attention, with extensive research focusing on two core processes: communication topology construction and model aggregation. D-PSGD~\cite{DPSGD} is one of the most commonly used methods in DFL, adopting a static ring communication topology and employing weighted averaging for model aggregation among participants. On the other hand, TornadoAggregate~\cite{lee2020tornadoaggregate} designs a static hybrid topology that combines star and ring topologies to further improve model performance, but experimental results demonstrate an increase in communication cost. However, such static-topology DFL methods fail to incorporate variations in training dynamics among participants during the learning process, thereby leading to performance variance in complex systems, particularly in the presence of system and data heterogeneity within EC systems. By contrast, dynamic-topology DFL methods offer enhanced adaptability by allowing the communication topology to adjust in response to complex conditions within EC systems. Zec \textit{et al.}~\cite{DAC} considered the impact of communication topology on model performance under the setting of non-IID data distribution across participants. It proposed a similarity-based sampling strategy that allows each participant to adaptively select neighboring participants for model transmission, and adopts the weight averaging method for model aggregation. To addressing the challenge of system heterogeneity, Wang \textit{et al.}~\cite{wang2020service} focused on optimizing topology construction between edge devices and edge servers with heterogeneous communication resources to minimize the total energy budget. Zeng \textit{et al.}~\cite{zeng2022gnn} proposed a cost-efficient topology optimization problem in a multi-tier edge environment and devised a scheduling algorithm to solve it, thereby minimizing the learning cost associated with graph neural network (GNN) processing across distributed edge servers with heterogeneous computation and communication capabilities.

Another line of work in DFL argues that naive model parameter averaging introduces variance in model updates during the aggregation process, thus focusing on designing an appropriate model aggregation mechanism. Valentina \textit{et al.}~\cite{zantedeschi2020fully} proposed a conditional model-updating algorithm based on Frank–Wolfe gradient descent, enabling participants to greedily update their models by incorporating the most important neighbor, adaptively identified by the algorithm. SGP~\cite{pushsum} utilizes the stochastic gradient push method, maintaining auxiliary variables to serve as anchors during the model aggregation process. On the other hand, to addressing the challenge of non-IID data, Jeon \textit{et al.}~\cite{jeon2021privacy} applied  alternating direction method of multiplier (ADMM) algorithm for decentralized model aggregation, but ADMM-based methods are constrained by the model consensus requirement.
Overall, prior studies have predominantly focused on either improving model performance or reducing energy consumption in isolation, addressing only parts of the aforementioned challenges in EC systems. Designing an efficient decentralized federated learning (DFL) framework, which entails constructing an optimal communication topology and selecting an effective model aggregation mechanism, remains an underexplored challenge. The key objective is to simultaneously maximize model performance and minimize cumulative energy costs throughout the collaborative learning process, particularly in EC environments where system and data heterogeneity pose significant hurdles.



\section{Preliminaries and Problem Formulation} 
\label{sec:3}
    
    

This section introduces the EC system structure and DFL basics, then identifies key optimization problems for deploying DFL within EC sytems.



\subsection{Edge Computing (EC) System}
\begin{figure}[t!]
\vspace{-3mm}
    \centering
    
    \includegraphics[width=1\linewidth]{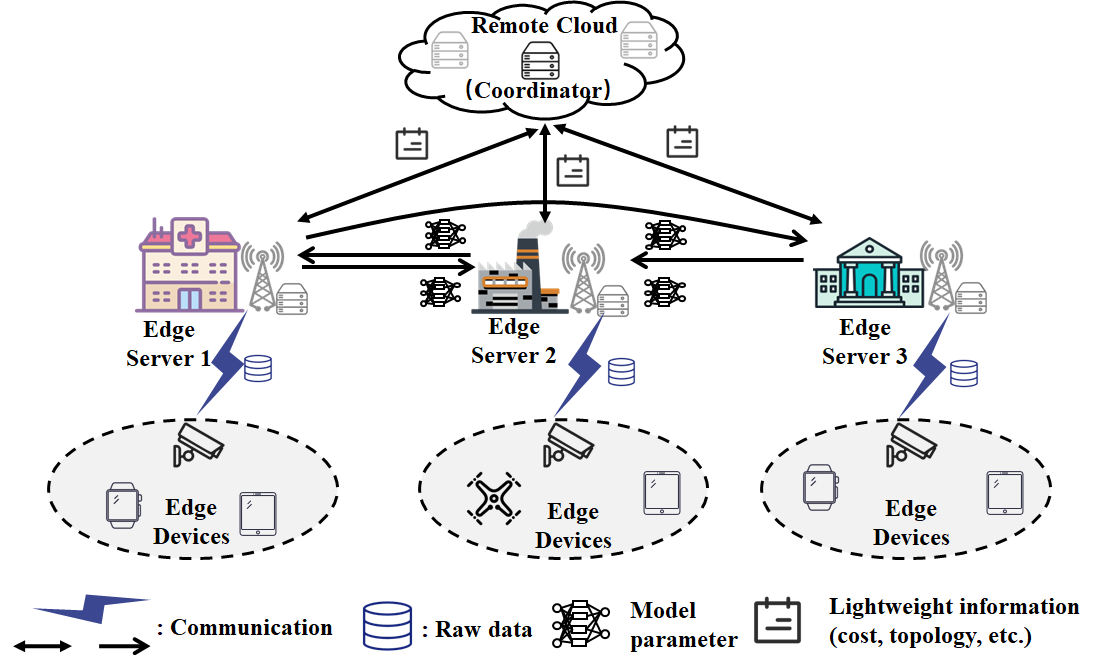}    
  \caption{The architecture of the edge computing (EC) system}
  \label{fig:1} 
  \vspace{-5mm}
\end{figure} 
 As illustrated in \figurename~\ref{fig:1}, we consider an EC system composed of a coordinator (\textit{i.e. a cloud server}) and a set of edge servers $\mathcal{S} = \{s_{i}\}_{i=1}^{N}$, where each server $s_i$ manages $|\mathcal{D}(s_i)|$ edge devices. Let $\mathcal{D}(s_i) = \{d_j^i\}$ represent the set of edge devices managed by edge server $s_i$. $\mathcal{D} = \{\mathcal{D}(s_i)\}_{i=1}^{N}$ denotes the set of all edge devices in the EC system with the total number of edge devices is $M$. Each edge device $d_j^i$ acts as a data provider, which transmits data $D(d_j^i)$ to its corresponding edge server $s_i$. 
The device-to-edge topology can be modeled as a device-to-edge communication graph $\mathcal{G}^d = \{\mathcal{S}, \mathcal{D}, \mathcal{E}^d\}$, where $\mathcal{E}^d = \{(s_i, d_j^i)|s_i\in\mathcal{S}, d_j^i\in\mathcal{D}(s_i),  \emph{A}^d(s_i,d_j^i)= 1\}$ denotes the set of established device-to-edge communication links. The topology information of $\mathcal{G}^d$ can be incorporated into a device-to-edge topology matrix $\emph{A}^d\in\{0,1\}^{N\times M}$, where $\emph{A}^d(s_i, d_j^i) = 1$ if edge device $d_j^i$ communicates with edge server $s_i$, and $\emph{A}^d(s_i, d_j^i) = 0$ otherwise. Edge servers are deployed across various geographical locations (e.g., hospitals, factories), exhibiting heterogeneous computation and communication capacities. Each edge server $s_i\in\mathcal{S}$ possesses a local dataset $D(s_i)$ collected from the edge devices it manages, and the number of this local dataset is denoted by $|D(s_i)|$. 
The inter-edge communication topology is represented by $\mathcal{G}^s = \{\mathcal{S}, \mathcal{E}^s\}$, where $\mathcal{E}^s = \{(s_i, s_j)|s_i,s_j\in\mathcal{S},s_i\neq s_j, \emph{A}^s(s_i, s_j) = 1\}$ denotes the set of established inter-edge communication links. We represent the communication connectivity of the inter-edge network $\mathcal{G}^s$ using a binary adjacency matrix $\emph{A}^s\in\{0,1\}^{N\times N}$, where $\emph{A}^s(s_i,s_j) = 1$ if $(s_i, s_j)\in\mathcal{E}^s$ and $\emph{A}^s(s_i,s_j) = 0$ otherwise.

In EC systems, a coordinator gathers global model training data and resource status, then broadcasts the inter-edge communication topology to all edge servers (see \figurename~\ref{fig:1}). The coordinator mainly exchanges lightweight control information with edge servers, while inter-server communication involves higher-volume model parameter transfers. 
It should be noted that the coordinator is fundamentally different from the central PS, which is specifically responsible for collecting model parameters from edge servers and updating the global model. Since the data exchanged between the coordinator and edge servers (e.g., 100–300 KB~\cite{lyu2019optimal}) is substantially smaller than typical model parameter transmissions, we assume the coordinator in the EC system does not face significant communication overhead.

    
    

\subsection{Decentralized Federated Learning (DFL)}
    
    
DFL is a promising architecture where participants perform local model training and iteratively refine their models through peer-to-peer parameter exchanges with neighboring nodes~\cite{chen2022decentralized}. 
We consider a set of participants $\mathcal{V} = \{v_1,...,v_n\}$, where $n$ is the number of participants. Each participant trains a local model $\emph{m}(v_i)$ on its local dataset $D(v_i)$, and the loss function is denoted as $f_i(\emph{m}(v_i);D(v_i))$. The objective function for traditional DFL is defined as below~\cite{DPSGD}:
\begin{equation}
    \min_{\mathcal{M}=\{\emph{m}(v_i),...,\emph{m}(v_n)\} }f(\mathcal{M}) = \frac{1}{n}\sum_{i=1}^{n} f_i(\emph{m}(v_i);D(v_i))
\end{equation}

Moreover, the decentralized communication topology can be modeled as a directed graph $\mathcal{G}_k = \{\mathcal{V},\mathcal{E}_k\}$ at the $k$-th communication round, where the edge set $\mathcal{E}_k = \{(v_i,v_j)|v_i,v_j\in\mathcal{V}, v_i \neq v_j, \emph{A}_k(v_i,v_j) = 1\}$ represents the established communication links between participants. The topology matrix $\emph{A}_k\in\{0,1\}^{n\times n}$ encompass all connectivity information within $\mathcal{G}_k$, where $\emph{A}_k(v_i,v_j) = 1$ if $(v_i,v_j)\in\mathcal{E}_k$ and $\emph{A}_k(v_i,v_j) = 0$ otherwise. 
In a directed communication topology, we denote the in-neighbor set and out-neighbor set of $v_i$ as $\mathcal{N}_k^{in}(v_i) = \{v_j|v_j\in\mathcal{V}, \emph{A}_k(v_i,v_j)=1\}$ and $\mathcal{N}_k^{out}(v_i) = \{v_j|v_j\in\mathcal{V}, \emph{A}_k(v_j,v_i)=1\}$, respectively. At the beginning of the $k$-th communication round, participant $v_i$ performs local training by:
\begin{equation}
    \emph{m}_k(v_i) = \Tilde{\emph{m}}_{k}(v_i) - \mu\nabla f_{i}(\Tilde{\emph{m}}_{k}(v_i);D_k(v_i))
    \label{eq:local_train}
\end{equation}
where $\mu$ represents the learning rate for local training. The local model parameters of $v_i$ before and after local training are separately denoted as $\Tilde{\emph{m}}_{k}(v_i)$ and $\emph{m}_{k}(v_i)$. Upon completing the local training process, each participant transmits local model to its out-neighbors and receives models from in-neighbors according to the communication topology $\mathcal{G}_k(\emph{A}_k)$. Then, the model of $v_i$ is further updated as follows:
\begin{equation}
    \Tilde{\emph{m}}_{k+1}(v_i) = f_{agg}(\emph{m}_k(v_i),\{\emph{m}_k(v_j)\}_{v_j\in\mathcal{N}_k^{in}(v_i)})
\end{equation}
where $f_{agg}(\cdot)$ represents the model aggregation mechanism.

\subsection{Energy Cost Model}
We define three types of energy costs associated with deploying DFL in an EC system, where edge servers $s_i\in\mathcal{S}$ in this system are treated as participants in DFL as follows:
\begin{enumerate}[label=\arabic*)]
\item \textit{Data Transmission Cost}:
At the beginning of the $k$-th communication round, edge devices transmit their collected data to the corresponding edge servers according to the device-to-edge topology $\mathcal{G}^d_k(\emph{A}_k^d)$. Let $|D_k(d_j^i)| = n_{tr}, d_j^i\in\mathcal{D}$ denote the fixed amount of data transmitted by each device in every round, and we use $\psi$ to represent the unit transmission cost per device. The data transmission cost for edge server $s_i$ is defined as follows: 
\begin{equation}
    E_k^{dt}(s_i) = \sum_{d_j^i\in\mathcal{D}(s_i)}\psi\cdot \emph{A}_k^d(s_i,d_j^i)
\end{equation}
where device-to-edge connectivity follows a Bernoulli distribution denoted by $\emph{A}^d_k(s_i,d^i_j)\sim Bern(\rho), s_i\in\mathcal{S}, d^i_j\in\mathcal{D}(s_i)$~\cite{guo2021towards}. The parameter $\rho$ represents the success probability of edge devices connecting to edge servers, which influences the degree of time-varying data heterogeneity.

\item \textit{Computation Cost:} Since model aggregation incurs significantly lower computation cost compared to local training~\cite{shi2024analysis}, we focus on estimating the computation cost of local training at edge servers. Let $\tau$ denote the unit computation cost required to process a single data sample during local training, which varies across edge servers due to system heterogeneity in computation resources. At the $k$-th communication round, the total computation cost incurred from local training is given by:
\begin{equation}
    E^{cp}_k(s_i) =  |D_k(s_i)|\cdot\tau_i
\end{equation}
where $|D_k(s_i)|$ denotes the size of local training dataset on edge server $s_i$.
\item \textit{Model Transmission Cost:} After local training, edge servers exchange their model parameters according to the inter-edge topology $\mathcal{G}_k^s(\emph{A}_k^s)$. Here, we consider only the energy consumption of model uploading, as the downlink typically offers significantly higher communication capacity, such as bandwidth and transmission power, compared to the uplink~\cite{chen2023energy}. Then, the model transmission cost for transmitting model $\emph{m}_k(v_j)$ to edge server $s_i$ can be calculated as:
\begin{equation}
    E^{mt}_k(s_i,s_j) = \sigma^{i,j}
\end{equation}
The unit cost for transmitting a model from $s_j$ to $s_i$ is denoted by $\sigma^{i,j}$, which varies due to heterogeneous resources across inter-edge communication links. 

\end{enumerate}

We further represent the total energy cost in $k$-th communication round as follows:
\begin{equation}
        E_k(\mathcal{G}_k^s(\emph{A}_k^s))\!=\!\!\!\sum_{s_i\in\mathcal{S}}\!\!E^{dt}_k\!(s_i)+\!\!\sum_{s_i\in\mathcal{S}}\!\!E^{cp}_k\!(s_i)+\!\!\!\!\!\sum_{s_i,s_j\in\mathcal{S}}\!\!\!\!\!E_k^{mt}\!(s_i,\!s_j)\cdot\emph{A}_k^s(s_i,\!s_j)
\end{equation}

\subsection{Optimization Objective }
Given the initial set of models $\mathcal{M}_0$ and the set of communication rounds $\mathcal{K} = \{1,...,K\}$, we aim to simultaneously maximize final model predictive accuracy and minimize the total energy consumption throughout the collaborative learning process. 
This is achieved by optimizing the inter-edge topology $\mathcal{G}_k^s(\emph{A}_k^s)$ across communication rounds and designing an effective model aggregation mechanism $f_{agg}$, which is formulated as the following multi-objective optimization problem:

{\small
\begin{equation}
\begin{split}
        &{\footnotesize \text{P1:}\!\!\!\min_{f_{agg},\{\mathcal{G}_k^s(\emph{A}_k^s)\}_{k\in\mathcal{K}}}\!\!\!\{\!-\mathbb{E}[F(\mathcal{M}_0, f_{agg},\{\mathcal{G}_k^s(\emph{A}_k^s)\}_{k\in\mathcal{K}})],\sum_{k\in\mathcal{K}}\!E_k\!(\mathcal{G}_k^s(\emph{A}_k^s)\!)\}}\\
    &\text{s.t.}\quad\emph{A}^s_{k}\in \{0,1\}^{N\times N},\quad k\in\mathcal{K}
\end{split}
\end{equation}}

\noindent Here, $\mathbb{E}[F(\mathcal{M}_0,f_{agg},\{\mathcal{G}_k^s(\emph{A}_k^d)\}_{k\in\mathcal{K}})]$ denotes the expected final predictive accuracy of the model across edge servers after completing the $K$-th round of the learning process. 

Solving Problem \text{P1} presents two key challenges: First, deriving an exact closed-form expression is critical for the multi-objective optimization problem considering the opposite optimization directions of the two objectives. While reducing inter-edge communication links decreases transmission costs, it may simultaneously degrade model performance during training. Second, both energy consumption and model predictive accuracy can only be measured after each training round, rendering them unavailable in advance~\cite{shi2024analysis}. Consequently, the dynamic inter-edge topologies must be determined without prior knowledge of these evaluation metrics, introducing significant complexity.

\section{Analysis and Transformation of the Optimization Objective}
\label{sec:4}
To solve Problem \text{P1}, we formulate an alternative problem that explicitly analyzes how the dynamics of topology construction influence model performance and energy consumption. Furthermore, we introduce a new metric, named utility, to quantify the impact of inter-edge connectivity on both energy cost and model performance. The metric denoted as:
\begin{equation}
    u_k^{i,j} = \alpha S^{p}_k(s_i,s_j) + (1-\alpha) S^{c}_k(s_i,s_j)
    \label{eq:ture utility}
\end{equation}
where $\alpha\in[0,1]$ is a weight hyperparameter that controls the trade-off between the normalized model performance factor $S_k^{p}$ and the energy cost factor $S_k^{c}$ within the utility evaluation. The normalized energy cost factor is defined as follows:
\begin{equation}
\begin{aligned}
    \Tilde{S}_k^c(s_i,s_j) &= E^{dt}_k\!(s_j)\!+\! E^{cp}_k\!(s_j)\!+\!E^{mt}_k\!(s_i,s_j) \\
     S_k^{c}(s_i,s_j) &= 1- Norm(\Tilde{S}_k^c(s_i,s_j)) 
\end{aligned}
\end{equation}

$\Tilde{S}_k^c(s_i,s_j)$ denotes the energy cost factor. $Norm(\cdot)$ denotes the normalization operation that maps a value to the range $[0, 1]$ by softmax function. Similarly, the normalized model performance factor is defined as:
\begin{equation}
\begin{aligned}
    \Tilde{S}_k^p(s_i,s_j) &= P_k^i-P_{k-1}^i \\
    S_k^{p}(s_i,s_j) &= Norm(\Tilde{S}_k^p(s_i,s_j))
    \label{eq:performance score}    
\end{aligned}
\end{equation}

The model performance factor $\Tilde{S}_k^p(s_i,s_j)$ evaluates model accuracy improvement at edge server $s_i$ between the $k$-th communication round and the previous round, while it is also scaled to the range $[0, 1]$ using softmax normalization.

Based on the definition of utility, we propose a tractable approximation of Problem \text{P1} that aims to maximize the total utility by optimizing the inter-edge topology across communication rounds, with the same objective of attaining the best model performance while minimizing cumulative energy consumption. 
The topology construction optimization problem \text{P2} is defined as follows:
\begin{equation}
\begin{split}
        &\text{P2:} \max_{\{\mathcal{G}_k^s(\emph{A}_k^s)\}_{k\in\mathcal{K}}}\sum_{k\in\mathcal{K}}\sum_{s_i,s_j\in S,s_i\neq s_j}\emph{A}_k^s(s_i,s_j)\cdot u_k^{i,j} \\
    &\text{s.t.}\quad\emph{A}_k^s(s_i,s_j) \in \{0,1\}^{N\times N},\quad s_i,s_j\in\mathcal{S},\quad k\in\mathcal{K} \\
   &\quad\quad\sum_{s_i,s_j\in\mathcal{S},s_i\neq s_j}\!\!\emph{A}_k^s(s_i,s_j) = \gamma\cdot|\mathcal{E}^s|,\quad k\in\mathcal{K}
\end{split}
\end{equation}
Here, $|\mathcal{E}^s|$ denotes the total number of possible inter-edge communication links, which is quadratic in the number of edge servers $N$. The hyperparameter $\gamma\in(0,1]$ controls the sparsity of the inter-edge topology, where decreasing $\gamma$ lead to fewer inter-edge communication links.

Optimizing \text{P2} is non-trivial due to its combinatorial complexity. As demonstrated in Theorem 1, the problem \text{P2} is NP-Hard, implying the absence of polynomial algorithm unless \text{P=NP}.
\begin{theorem}
The topology construction problem \text{P2}, which aims to maximize total utility under a constraint on the number of communication links, is NP-hard.
\end{theorem}
\quad\textit{Proof:}
We prove the NP-hardness of \text{P2} through a polynomial-time reduction from the Zero-One knapsack (ZOK) problem, which is known to be NP-hard~\cite{kellerer2004introduction}. Formally, given an item set $\mathbb{L} = \{l\}$, where each item $l$ has weight $e_{l}$ and value $v_{l}$, and a knapsack with capacity $C$, the ZOK problem aims to select a subset of items $\mathbb{L}^{*}\subseteq\mathbb{L}$ that maximizes the total value, while ensuring that the total weight does not exceed the given capacity $C$. Consider an instance of \text{P2}, denoted by $I_{\textbf{P2}}$, where the number of communication rounds $K$ is set to 1. We construct a polynomial reduction by mapping the set of inter-edge communication links $\{(s_i,s_j)|s_i,s_j\in\mathcal{S}, s_i\neq s_j\}$ to the item set $\mathbb{L}$. The utility $u_{k}^{i,j}$ of each inter-edge communication link $(s_i,s_j)$ is interpreted as the value of an item, with the item weight set to $w_{l} = 1, \forall l\in\mathbb{L}$. The constrained number of inter-edge communication links, given by $\gamma\cdot|\mathcal{E}^s|$, is treated as the knapsack capacity $C$. Since the constraints in $I_{\textbf{P2}}$ align with the setting of knapsack problem, $I_{\textbf{P2}}$ is exactly a ZOK problem. Given the NP-hardness of the ZOK problem, the constructed polynomial reduction indicates that $P2$ is also NP-hard, thus completing the proof.

The utility of inter-edge communication links may vary over communication rounds. For example, variations in local training across edge servers can affect the performance of the aggregated model, resulting in dynamic changes in link utility. Consequently, we do not assume that utility values are drawn from any particular distribution. Instead, utility must be estimated prior to each communication round without relying on statistical assumptions. 

However, directly solving the problem under such uncertain and dynamic conditions is infeasible with conventional dynamic programming or greedy algorithms designed for the 0-1 knapsack (ZOK) problem~\cite{goel2006asking}. To address this challenge, we reformulate the problem within the multi-armed adversarial bandit (MAB) framework, which offers a lightweight decision-making scheme and provides well-established theoretical performance guarantees~\cite{lattimore2020bandit}. 
In the MAB problem, each inter-edge communication link can be treated as an arm, and its utility is regarded as the reward associated with that arm. The reward is assumed to be a predetermined value, but it remains unknown until the corresponding inter-edge communication link is selected to construct the communication topology across communication rounds. Inspired by the MAB problem, which aims to maximize the expected cumulative reward, we reformulate Problem \text{P2} to maximize the expected total utility across communication rounds, as follows:
\begin{equation}
\begin{split}
        &\text{P3:} \max_{\{\mathcal{G}_k^s(\emph{A}_k^s)\}_{k\in\mathcal{K}}}\mathbb{E}\bigg[\sum_{k\in\mathcal{K}}\sum_{s_i,s_j\in\mathcal{S},s_i\neq s_j}\emph{A}_k^s(s_i,s_j)\cdot u_k^{i,j}\bigg] \\
    &\text{s.t.}\quad\emph{A}^s_{k}(s_i,s_j) \in \{0,1\}^{n\times N},\quad s_i,s_j\in\mathcal{S},\quad k\in\mathcal{K} \\
   &\quad\quad\sum_{s_i,s_j\in\mathcal{S},s_i\neq s_j}\!\!\emph{A}_k^s(s_i,s_j) = \gamma\cdot|\mathcal{E}^s|,\quad k\in\mathcal{K}
\end{split}
\end{equation}
where $\emph{A}_k^s$ is a set of random variables. Let $\mathbb{E}[\mathbb{I}\{\emph{A}_k^s(s_i,s_j) = 1\}] = p_k(s_i,s_j)$ represent the expected probability of selecting the inter-edge communication link $(s_i,s_j)$ at the $k$-th communication round, where $\mathbb{I}\{\cdot\}$ denotes the indicator function. Thus, the problem of optimizing the inter-edge topology can be transformed into a probability assignment problem over the set of inter-edge communication links, which is defined as follows: 
\begin{equation}
\begin{split}
        &\text{P4:} \max_{\{\emph{p}_k\}_{k\in\mathcal{K}}}\sum_{k\in\mathcal{K}}\sum_{s_i,s_j\in\mathcal{S}}p_k(s_i,s_j)\cdot u_k^{i,j} \\
    &\text{s.t.}\quad p_k(s_i,s_j)\in[0,1],\!\!\quad s_i,s_j\in\mathcal{S},\!\!\quad s_i\neq s_j,\!\!\quad k\in\mathcal{K}\\
    &\quad\sum_{s_i,s_j\in\mathcal{S},s_i\neq s_j}p_k(s_i,s_j) = \gamma\cdot|\mathcal{E}^s|,\quad k\in\mathcal{K}
\end{split}
\end{equation}
where $\emph{p}_k = \{p_k(s_i,s_j)|s_i,s_j\in\mathcal{S}, s_i\neq s_j\}$ denotes the set of probabilities assigned to the corresponding inter-edge communication links at the $k$-th communication round. 

In summary, the transformation from \text{P1} to \text{P4} proceeds through several key steps. First, we introduce a new evaluation metric named utility to approximate the multi-objectives optimization problem \text{P1} with an alternative problem, \text{P2}. Given the NP-hardness of \text{P2} and the inherent uncertainty and dynamism of the setting, we reformulate \text{P2} into \text{P3}, aiming to maximize the expected total utility across communication rounds, inspired by the MAB problem. Finally, for facilitating analysis, the selection of a subset of inter-edge communication links for topology construction is transformed into a probability assignment problem over the link set. 

\begin{figure*}[t!]
\vspace{-2mm}
    \centering
    
    \includegraphics[width=0.86\linewidth]{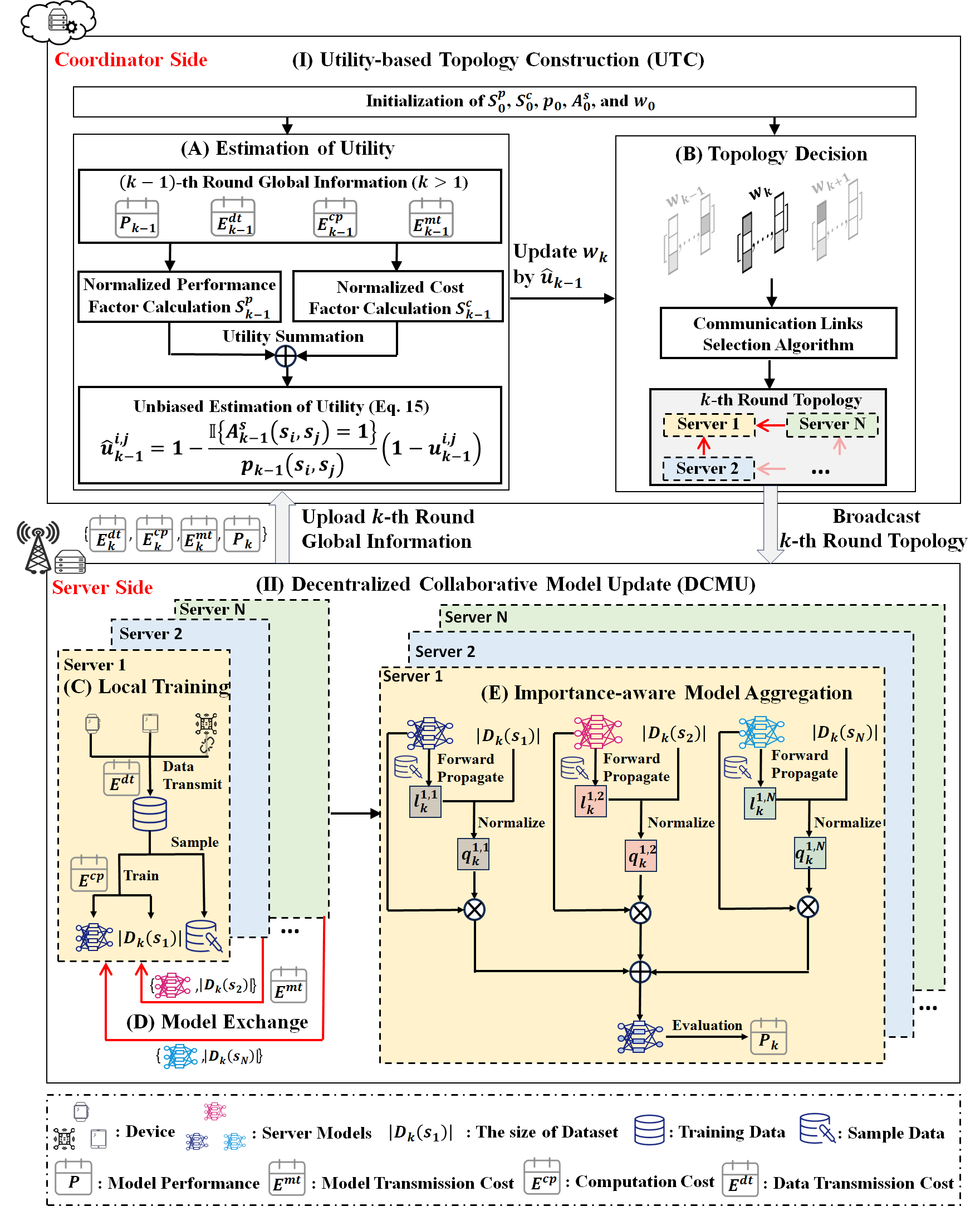}
       
  \caption{The overview of our proposed algorithm.}
  \label{fig:overview} 
  \vspace{-5mm}
\end{figure*}

\section{Algorithm Design}
\label{sec:5}
In this section, we present the proposed two-phase algorithm, which consists of Utility-based Topology Construction (UTC) and Decentralized Collaborative Model Update (DCMU), designed to solve the joint optimization problem for deploying DFL in heterogeneous EC systems, as illustrated in \figurename~\ref{fig:overview}. In phase \uppercase\expandafter{\romannumeral 1}, UTC (Algorithm~\ref{algo:2}) constructs the optimal inter-edge topology using the communication links selection algorithm (Algorithm~\ref{algo:4}), based on an unbiased estimation of the observed utility associated with inter-edge communication links. This process is executed on the coordinator side across communication rounds. In phase \uppercase\expandafter{\romannumeral 2}, DCMU (Algorithm~\ref{algo:3}) is performed on the edge server side. Each edge server trains locally, exchanges model parameters based on the constructed topology, and updates its models using an importance-aware model aggregation mechanism to mitigate the degradation in model performance caused by data heterogeneity. The overall procedure is summarized in Algorithm~\ref{algo:1}.
\renewcommand{\thealgorithm}{1}
\begin{algorithm}
\caption{The Two-phase Algorithm }\label{algo:1}
\begin{algorithmic}[1]
\Require The set of edge servers $\mathcal{S}$, the set of edge devices $\mathcal{D}$, the number of communication round $K$, the number of communication links to be selected $\gamma\cdot|\mathcal{E}^s|$ and the device-to-edge topology $\{\mathcal{G}^d_k(\emph{A}^d_k)\}_{k=1}^{K}$
\State Initialize $\emph{m}_0(s_i)$ at each edge server, and $S_0^p$, $S_0^c$, $\emph{A}_0^s$, $\emph{w}_0$, and $\emph{p}_0$ at the coordinator
\For{each round $k\in [1,2,...,K]$}
\State Phase \uppercase\expandafter{\romannumeral 1}: run Algorithm~\ref{algo:2} to determine the optimal inter-edge topology $\mathcal{G}_k^s(\emph{A}_k^s)$ and broadcast it to edge serves \Comment{on the coordinator side}
\State Phase \uppercase\expandafter{\romannumeral 2}: run Algorithm~\ref{algo:3} to perform collaborative model updates \Comment{on the edge server side}
\EndFor
\end{algorithmic}
\end{algorithm}

\subsection{Phase \uppercase\expandafter{\romannumeral 1}: Utility-based Topology Construction (UTC)}
As previously discussed, we formulate an optimization problem to design the inter-edge topology based on a utility metric. However, the utility of inter-edge communication links in each communication round remains unknown prior to the completion of the learning process in that round. To address this challenge, we propose UTC, which determines the optimal inter-edge topology for the current round using an unbiased estimation of utility derived from information collected in the previous round.
\renewcommand{\thealgorithm}{2}
\begin{algorithm}
\caption{Utility-based Topology Construction (UTC)}\label{algo:2}
\begin{algorithmic}[1]
\Require The number of communication links to be selected $\gamma\cdot|\mathcal{E}^s|$, the selection weight $\{w_{k-1}(s_i,s_j)\}_{s_i,s_j\in\mathcal{S}, s_i\neq s_j}$, the set of selection probabilities $\emph{p}_{k-1}$, and the topology matrix $\emph{A}_{k-1}^s$
\Ensure The inter-edge topology $\mathcal{G}_k^{s}(\emph{A}_k^s)$
\For{each edge server $s_i\in\mathcal{S}$}
\For{each edge server $s_j\in\{\mathcal{S}\backslash\{s_i\}\}$}
\State $u^{i,j}_{k-1}\leftarrow$ Calculate the true utility using Eq.~\ref{eq:ture utility}
\State $\hat{u}^{i,j}_{k-1}\leftarrow$ Unbiased estimation of utility by Eq.~\ref{eq:unbiased}
\State $w_{k}(s_i,s_j)\leftarrow$ Update selection weight by Eq.\ref{eq:w_update}
\EndFor
\EndFor
\State Determine inter-edge topology $\mathcal{G}_k^s(\emph{A}_k^s)$ through $\text{CLinkSec}(\{w_k(s_i,s_j)\}_{s_i,s_j\in\mathcal{S}, s_i\neq s_j},\gamma\cdot|\mathcal{E}^s|)$\State Return $\mathcal{G}_k^{s}(\emph{A}_k^s)$
\end{algorithmic}
\end{algorithm}

At the beginning of the $k$-th communication round, the utility of the communication links from the previous round is computed based on the uploaded global information, including model predictive accuracy and the three types of energy cost, as defined in Eq.~\ref{eq:ture utility} to Eq.~\ref{eq:performance score}. The crucial step in UTC is accurately estimating the utility based on the observed true utility from the $(k-1)$-th communication round to guide the communication topology decision in the $k$-th round, as demonstrated in the part (A) in \figurename~\ref{fig:overview}. We use an unbiased estimation $\hat{u}^{i,j}_{k-1}$, which is defined as follows:
\begin{equation}
    \hat{u}^{i,j}_{k-1} = 1-\frac{\mathbb{I}\{\emph{A}^{s}_{k-1}(s_i,s_j) = 1\}}{p_{k-1}(s_i,s_j)}(1-u^{i,j}_{k-1})
    \label{eq:unbiased}
\end{equation}
$\mathbb{I}\{\emph{A}^{s}_{k-1}(s_i,s_j)= 1\}$ represents whether inter-edge communication link $(s_i,s_j)$ is selected to construct topology in the $(k{-}1)$-th communication round. In the initialization process, the normalized model performance factor $S_0^p(s_i,s_j), s_i,s_j\in\mathcal{S}, s_i \neq s_j$ and energy cost factor $S_0^c(s_i,s_j), s_i,s_j\in\mathcal{S}, s_i \neq s_j$ for inter-edge communication links are both initialized to $0$, each element in the set of probabilities $\emph{p}_0$ is set to $1$, and the topology matrix is set to zero matrix $\emph{A}_0^s\in\{0\}^{N\times N}$

From Eq.~\ref{eq:unbiased}, it follows that $\mathbb{E}[\hat{u}^{i,j}_{k-1}] = u^{i,j}_{k-1}$, indicating that $\hat{u}^{i,j}_{k-1}$ serves as an unbiased estimator of $u^{i,j}_{k-1}$, with the proof provided in the Appendix A. The unbiased estimator accurately reveals the potential utility of inter-edge communication links, after which an exponential selection weight is applied to map this utility to the importance of selecting the corresponding link. The exponential selection weight for each inter-edge communication link $(s_i,s_j)$ is updated as follows:
\begin{equation}
    w_k(s_i,s_j) = w_{k-1}(s_i,s_j)\exp{(\eta\hat{u}^{i,j}_{k-1})}
    \label{eq:w_update}
\end{equation}
where $w_k(s_i,s_j)$ denotes the selection weight of the inter-edge communication link $(s_i,s_j)$ in $k$-th communication round. 
Specifically, $w_0(s_i,s_j), s_i,s_j\in\mathcal{S}, s_i \neq s_j$ is set to $0$ during the initialization process. The parameter $\eta\in(0,1]$ serves as a tuning parameter that balances exploration and exploitation. A smaller value of $\eta$ encourages UTC to explore previously unselected inter-edge communication links, while a larger $\eta$ biases it toward exploiting known high-utility links. 

Given the set of selection weights $\{w_k(s_i,s_j)\}_{s_i,s_j\in\mathcal{S}, s_i\neq s_j}$ and the required number of links to be selected $\gamma\cdot|\mathcal{E}^s|$, we employ a communication link selection algorithm (Algorithm~\ref{algo:4}) to assign probabilities to inter-edge communication links. The algorithm first updates the selection probability based on Eq.~\ref{eq:prob}, and then applies a capping function to constrain each probability to a maximum of 1 (line 4 in Algorithm~\ref{algo:4}), as the exponential selection weights may otherwise become excessively large and result in probabilities exceeding 1.
For the selection of communication link, we apply a stochastic selection method based on dependent rounding algorithm~\cite{gandhi2006dependent} to select the inter-edge communication links to construct the topology (line 7 to line 19 in Algorithm~\ref{algo:4}). Specifically, the selection probabilities for inter-edge links are continuously updated according to Eq.~\ref{eq:dependent} (line 16 in Algorithm~\ref{algo:4}), ensuring that their expected values remain unchanged and thereby preserving the optimal probability assignment for maximizing total utility, as formally proven in Appendix B.


The output of Algorithm~\ref{algo:4} is the subset of communication links used to construct the inter-edge topology in the $k$-th communication round. Each element in inter-edge topology matrix $\emph{A}_k^s(s_i,s_j)$ is assigned a value of $1$ if $(s_i,s_j)\in \mathcal{E}_k^s$, and a value of $0$ otherwise. After that, the coordinator broadcasts $k$-th inter-edge topology to edge servers.
\renewcommand{\thealgorithm}{3}
\begin{algorithm}
\caption{Decentralized Collaborative Model Update (DCMU)}\label{algo:3}
\begin{algorithmic}[1]
\Require The device-to-edge topology $\mathcal{G}^d_k(\emph{A}^d_k)$ and the inter-edge topology $\mathcal{G}_k^s(\emph{A}_k^s)$
\For{each edge server $s_i\in\mathcal{S}$}
\State Collect data from edge devices $d_j^i\in\mathcal{D}(s_i)$ based on $\mathcal{G}^d_k(\emph{A}^d_k)$ and record data transmission cost $E^{dt}_k$
\State Local training (Eq.~\ref{eq:local_train}) and record computation cost $E^{cp}_k$
\State Record the size of the training dataset $|D_k(s_i)|$ and construct a sampled training dataset $D^{sa}_k(s_i)$
\State Model exchanges based on $\mathcal{G}_k^s(\emph{A}_k^s)$ and record model transmission cost $E_k^{mt}$
\State Calculate the approximated importance of models from in-neighboring server
\State Assign aggregation weights by Eq.~\ref{eq:agg_w_assign}
\State Perform importance-aware model aggregation (Eq.~\ref{eq:agg})
\State Evaluate model predictive accuracy and record the result $P_k$
\State Upload global information to the coordinator, including $E_k^{dt}$, $E_k^{cp}$, $E_k^{mt}$ and $P_k$
\EndFor
\end{algorithmic}
\end{algorithm}

\renewcommand{\thealgorithm}{4}
\begin{algorithm}
\caption{Communication Links Selection (CLinkSec)}\label{algo:4}
\begin{algorithmic}[1]
\Require The selection weight $\{w_k(s_i,s_j)\}_{s_i,s_j\in\mathcal{S}, s_i\neq s_j}$ and the number of communication links to be selected $\gamma\cdot|\mathcal{E}^s|$
\Ensure The selected set of communication links $\hat{\mathcal{E}}^s_k$
\For{each edge server $s_i\in\mathcal{S}$}
\For{each edge server $s_j\in\{\mathcal{S}\backslash\{s_i\}\}$}
\State\begin{equation}
    \hat{p}_k(s_i,s_j) = \frac{\gamma\cdot|\mathcal{E}^s|\cdot w_k(s_i,s_j)}{\sum_{s_i,s_j\in\mathcal{S},s_i\neq s_j}w_k(s_i,s_j)}
    \label{eq:prob}
\end{equation}
\State\quad$p_k(s_i,s_j) = \min{(\hat{p}_k(s_i,s_j),1)}$
\EndFor
\EndFor
\While{$\text{len}(\{0<p_k(s_i,s_j)<1\}_{s_i,s_j\in\mathcal{S},s_i\neq s_j})>0$}
\If{$\text{len}(\{0<p_k(s_i,s_j)<1\}_{s_i,s_j\in\mathcal{S},s_i\neq s_j}) = 1$}
\If{$p_k(s_i,s_j)<0.5$}
\State $p_k(s_i,s_j) = 0$
\Else
\State $p_k(s_i,s_j) =1 $
\EndIf
\Else
\State Randomly select two probabilities $p_i$ and $p_j$ from $\emph{p}_k = \{p_k(s_i,s_j)\}_{s_i,s_j\in\mathcal{S}, s_i\neq s_j}$, and let $\theta = \min{(1-p_i, p_j)}$ and $\delta = \min{(p_i, 1-p_j})$.
\State Update $p_i$ and $p_j$ by
\begin{equation}
\begin{aligned}
(p_i,p_j)&=
&\begin{cases}
(p_i+\theta,p_j-\theta)& \text{with probability}\quad \frac{\delta}{\theta+\delta},\\
(p_i-\delta,p_j+\delta)& \text{with probability} \quad\frac{\theta}{\theta+\delta}
\end{cases}
\end{aligned}
\label{eq:dependent}
\end{equation}
\EndIf
\EndWhile
\State $\mathcal{E}_k^{s} = \{(s_i,s_j):p^s_k(s_i,s_j)=1\}_{s_i,s_j\in\mathcal{S}, s_i\neq s_j}$
\State Return $\mathcal{E}_k^{s}$
\end{algorithmic}
\end{algorithm}


\subsection{Phase \uppercase\expandafter{\romannumeral 2}: Decentralized Collaborative Model Update (DCMU)}
After the inter-edge topology is determined, DCMU (Algorithm~\ref{algo:3}) is executed on the edge server side to efficiently train local models. Each communication round involves local training, model exchange, and importance-aware model aggregation, as illustrated in the lower part of \figurename~\ref{fig:overview}.
\subsubsection{Local Training and Model Exchange}
Edge devices first upload their collected data $D_k(d_j^i)$ to the corresponding edge servers, based on the device-to-edge topology $\mathcal{G}_k^d(\emph{A}_k^d)$, to construct the local training dataset on edge servers, denoted as $D_k(s_i) = \bigcup_{d^i_j\in\mathcal{D}(s_i), \emph{A}^d_k(s_i,d^i_j) = 1}D_k(d^i_j)$. Then, edge servers update the local models on their training data $D_k(s_i), s_i\in\mathcal{S}$, as denoted in Eq.~\ref{eq:local_train}. During this process, each edge server randomly samples a subset of its training data, denoted as $D_k^{sa}(s_i)$, with a fixed size $B$, and records the size of its full training dataset $|D_k(s_i)|$. These sets of sample data and recorded size information are subsequently used in the model aggregation step. 

Upon completing local training, each edge server $s_i$ transmits its local model and recorded size information to the servers in its out-neighbor set $\mathcal{N}^{out}_{k}(s_i)$, while receiving the corresponding information from those in its in-neighbor set $\mathcal{N}^{in}_{k}(s_i)$, as illustrated in the part (D) in \figurename~\ref{fig:overview}.



\subsubsection{Importance-aware Model Aggregation}
A key challenge in EC systems is the data heterogeneity encountered during the learning process, manifested as non-IID and time-varying training data. Such heterogeneity biases local model training toward divergent convergence directions, significantly undermining the effectiveness of model aggregation, as conventional simple averaging tends to amplify this divergence~\cite{xu2024overcoming,nascimento2024data,cui2023data,zheng2024adaptive}. 
To address this challenge, we propose an importance-aware model aggregation mechanism that evaluates the importance of models received from in-neighboring edge servers. 

Prior works~\cite{lai2021oort,liu2024fedasmu} have demonstrated that the optimal weight assignment in the model aggregation process is proportional to the importance of each model, which can be evaluated by testing the models on a sampled training dataset, denoted as $|D^{sa}|\cdot\sqrt{\frac{1}{|D^{sa}|}\sum_{\xi\in D^{sa}}\|\nabla f(\emph{m}, \xi)\|^2)}$. Specifically, $\|\nabla f(\emph{m}, \xi)\|$ denotes the L2-norm of the model $\emph{m}$'s gradient, computed on a sample dataset $D^{sa}$, where $\emph{m}$ is one of the models involved in the aggregation process. Intuitively, a model with a larger gradient norm is considered more important and is thus assigned a greater aggregation weight~\cite{lai2021oort}. However, computing gradient norms for models on the sampled data requires additional computational overhead. To reduce this cost, we approximate model importance at edge server $s_i$ by calculating the loss of model $\emph{m}_k(s_j), s_j\in\mathcal{N}_k^{in}(s_i)\cup\{s_i\}$ on the sample data $D_k^{sa}(s_i)$, based on the fact that larger gradient norms often correspond to higher loss values~\cite{johnson2018training}. 
The approximate importance of the model $\emph{m}_k(s_j)$ at edge server $s_i$ can be defined as follows:
\begin{equation}
\begin{split}
        &loss_k^{i,j}(\xi) = f_{\xi\sim D_k^{sa}(s_i)}(m_{k}(s_j), \xi)\\
        &l_k^{i,j} = B\cdot\sqrt{\frac{1}{B}\sum_{\xi\in D_k^{sa}(s_i)}(loss_k^{i,j}(\xi))^2}
\end{split}
\end{equation}
where $\xi\sim D_k^{sa}(s_i)$ is drawn from the sample data. Local loss function $f(\cdot,\cdot)$ only involves the forward propagation progress, and the size of sample data $B$ is much smaller than that of the training dataset. Therefore, the computational cost of this process can be negligible. 

Moreover, due to the randomness in the device-to-edge topology, the size of training dataset among edge severs varies across communication rounds. Models trained on smaller datasets tend to exhibit greater variance and are therefore assigned smaller aggregation weights. Finally, the aggregation weight for updating the model on the edge server $s_i$, incorporating both the approximated model importance and the influence of the model’s training dataset size, is calculated as follows:
\begin{equation}
    q^{i,j}_k = \beta\cdot\text{Softmax}(l_k^{i,j}) + (1-\beta)\cdot\text{Softmax}(|D_k^{s_j}|)
    \label{eq:agg_w_assign}
\end{equation}
where $\text{Softmax}(\cdot)$ denotes the softmax function used to normalization, and the balance hyperparameter $\beta\in[0,1]$ controls the trade-off between the effects of the approximated model importance and the training dataset size on model aggregation. 

After the aggregation weights are assigned, each edge server $s_i$ updates its local model $\emph{m}_{k}(s_i)$ by aggregating the models $\emph{m}_k(s_j), s_j\in\mathcal{N}_k^{in}(s_i)\cup\{s_i\}$, as denoted as follows:
\begin{equation}
    \hat{\emph{m}}_{k+1}(s_i) = \sum_{s_j\in\mathcal{N}_k^{in}(s_i)\cup\{s_i\}}q^{i,j}_k\cdot\emph{m}_k(s_j)
    \label{eq:agg}
\end{equation}

Upon completing the importance-aware model aggregation process, each edge server evaluates the predictive accuracy of its local model and records the result $P_k$. 
Moreover, edge servers also log the data transmission cost $E^{dt}_k$, computation cost $E^{cp}_k$ and model transmission cost $E^{mt}_k$ during the respective stages in DEMU. This recorded information, collectively regarded as the global information for the $k$-th communication round, is uploaded to the coordinator to guide the construction of the inter-edge topology for the subsequent round, as indicated by the grey arrow in the left part of \figurename~\ref{fig:overview}.

\subsection{Theoretical Analysis}
 
To demonstrate the effectiveness of two-stage algorithm, we provide a theoretical regret guarantee. To facilitate the proof, we first introduce the following definition.
\begin{definition}
    Let $p^*_{k}(s_i,s_j)$ represent the optimal probability assigned to the inter-edge communication link $(s_i,s_j)$ in $k$-th communication round. The expected cumulative utility obtained under the optimal inter-edge topology is defined as follows: 
    \begin{equation}
        G^{*}_K = \sum_{k\in\mathcal{K}}\sum_{s_i,s_j\in\mathcal{S},s_i\neq s_j}p^{*}_{k}(i,j)\cdot u_{k}^{i,j}
    \end{equation}
\end{definition}
\begin{definition}
    Given the optimal cumulative utility $ G^{*}_K$, the regret of the proposed algorithm is defined as follows:
    \begin{equation}
        R_{K} =  G^*_K- \sum_{k\in\mathcal{K}}\sum_{s_i,s_j\in\mathcal{S},s_i\neq s_j}p_{k}(i,j)\cdot u_{k}^{i,j}
    \end{equation}
\end{definition}

Intuitively, regret quantifies the performance gap between the proposed algorithm and the optimal solution, serving as a key metric for evaluating the effectiveness of the algorithm. Based on Definition 1 and 2, we present the following theorem, which establishes an upper bound on the regret.
\begin{theorem}
    The regret of \pname can be bounded as 
    \begin{equation}
        R_K \leq \eta KN^2 + \frac{2\log{N}}{\eta},
      \end{equation}
    and if the tuning parameter $\eta = \frac{\sqrt{K\log{N}}}{NK}$, we have
    \begin{equation}
        R_K \leq 3N\sqrt{K\log{N}}.
    \end{equation}
\end{theorem}
\quad\textit{Proof:} The complete proof has been presented in Appendix C.

Theorem 2 demonstrates that our proposed algorithm achieves a constant regret upper bound compared to the optimal solution.

We further analyze the time complexity of the proposed algorithm. In the UTC phase, utility estimation incurs a time complexity of $\mathcal{O}(KN^2)$. The communication link selection algorithm is executed on edge servers during each communication round, where the \text{WHILE} loop (lines 7 to 18 in Algorithm~\ref{algo:4}) runs at most $N$ times, resulting in a time complexity of $\mathcal{O}(KN^2)$. Additionally, the computation of aggregation weights primarily involves evaluating the loss of each transmitted model on the sampled data stored at each server, with a time complexity of $\mathcal{O}(\phi BKN^2)$. Here, $\mathcal{O}(\phi)$ denotes the time complexity of computing the loss on a single data point, which is treated as a constant. Therefore, the overall time complexity of the proposed algorithm is $\mathcal{O}((\phi BK + 2K)N^2)$

\section{Experimental Study}
\label{sec:6}
In this section, we conduct extensive simulation experiments to evaluate the effectiveness of our proposed solution from different perspectives. We aim to answer the following research questions:
\begin{itemize}
    \item \textbf{RQ1:} How does our proposed \pname perform in terms of model performance and energy cost compared to the four baselines?
    \item \textbf{RQ2:} How does the sparsity of the inter-edge communication topology $\gamma$ in UTC affect the final results?
    \item \textbf{RQ3:} How does the scalability of the proposed \pname vary with the number of edge servers in the EC system?
    \item \textbf{RQ4:} What is the impact of key hyperparameters $\alpha$ and $\beta$ on the \pname performance?
    \item \textbf{RQ5:} How do different components, such as utility-based topology construction and importance-aware model aggregation affect the performance of \pname?
\end{itemize}
\subsection{Experiment Setup}
\subsubsection{EC System Setup and Task Description} 
For our simulations, we first consider an EC system composed of $N=5$ edge servers and one coordinator, and each server manages $|\mathcal{D}(s_i)| = 30, s_i\in\mathcal{S}$ edge devices. To evaluate the proposed framework's performance, we conduct experiments on the image classification tasks using two datasets: Fashion-MNIST~\cite{xiao2017fashion} and CIFAR-10~\cite{krizhevsky2009learning}. Concretely, Fashion-MNIST consists of 60,000 grayscale images of size $28*28$, categorized into 10 classes, while CIFAR-10 comprises 60,000 color images of size $32*32$, also divided into 10 classes. Each pixel in every data sample from both datasets is stored as an 8-bit integer, implying that the size of each sample in Fashion-MNIST and CIFAR-10 is $BPS_{\text{fmnist}} = 28*28*1*8$ bits and $BPS_{\text{cifar}}=32*32*3*8$ bits, respectively. For both tasks, each edge server trains and evaluates a model comprising two convolution layers, two fully connected layers, and a output layer, encompassing $1,474,416$ training parameters for Fashion-MNIST and $4,224,288$ for CIFAR-10~\cite{DAC,shi2024analysis}. Each model parameter is stored as a 32-bit float. Accordingly, the total model size is $BPM_{\text{fmnist}} = 47.2$M bits and $BPM_{\text{cifar}} = 135.2$M bits for Fashion-MNIST and CIFAR-10, respectively.

\subsubsection{Simulation of Resources Heterogeneity at Edge}
To simulate heterogeneous edge computing resources, two types of machines, labeled as Type A (weak) and Type B (powerful), are assumed to be randomly deployed as edge servers in the EC system. The Type A machine is equipped with a 2.3GHz 4-Core Intel i5-6500TE processor and 4GB of RAM, whereas the Type B machine is configured with a 3.8GHz 16-Core Intel i9-12900F processor and 32GB of RAM. To gauge the unit computation cost $\tau_i$, we record the execution time on both types of machines by running the models on a single data sample from two datasets, repeating the process 100 times to obtain stable results. We then compute the energy consumption (in joules) for processing a single data sample based on the execution time and server power. To emulate heterogeneous edge communication resources, we assume that each inter-edge communication link is characterized by a distinct level of energy efficiency (EE), as proposed by the European Telecommunications Standards Institute (ETSI)~\cite{standard2015environmental}. EE is measured in kilobit/Joule (Kbit/J)~\cite{bjornson2018energy}. Generally. a higher EE indicates more efficient communication in EC systems, whereas a lower EE value reflects an energy-inefficient communication environment. Each inter-edge communication link is assumed to be randomly assigned an energy efficiency value $EE^{i,j}$ in the range of 20–50 Kbit/J~\cite{barbieri2023carbon}. The unit cost $\sigma^{i,j}$ for edge server $s_j$ to transmit its model to edge server $s_i$ is computed by $\frac{BPM}{EE^{i,j}}$. Similarly, the data transmission cost for transmitting data by a single device is denoted as $\frac{n_{tr}\times BPS}{EE^d}$, where the energy efficiency of device-to-edge communication $EE^d$ is set to 1 Kbit/J~\cite{barbieri2025close}.
\subsubsection{Training Settings}
The Fashion-MNIST and CIFAR-10 datasets are both split into 50,000 training images and 10,000 test images. To emulate non-IID data heterogeneity in the training dataset, we sample the label allocation ratio vector from a Dirichlet distribution for each edge server $s_i\in\mathcal{S}$, denoted as $\emph{r}_i\sim Dir(\lambda)$, where each element $r_i(j)$ represents the proportion of training images corresponding to a specific label type $j,j\in[1,...,|lb|]$~\cite{li2022federated}. $|lb|$ represents the number of distinct labels in the dataset, and $\lambda$ controls the degree of data distribution skewness (DDS). Specifically, when $\lambda\rightarrow 0$, the training data on each edge server is biased toward a single random label (\textit{i.e.}, high degree of non-IID).


In this experiment, the training dataset size $|D^{tr}(s_i)|$ for each edge server $s_i\in\mathcal{S}$ is set to 800, and the number of training samples $|D^{tr}_k(s_i)|$ used in each communication round $k\in\mathcal{K}$ is set to 60. To simulate time-varying data heterogeneity during the learning process, we randomly partition the set of training samples $D^{tr}_k(s_i),s_i\in\mathcal{S}$ into 30 subsets in each communication round, from which edge server $s_i$ randomly selects a number of subsets for local training. The number of selected subsets determined by device-to-edge topology $\mathcal{G}_k^d(\emph{A}_k^d)$. As previously discussed, device-to-edge connectivity follows a Bernoulli distribution $Bern(\rho)$. The parameter $\rho$ denotes the success probability of communication (SPC) between an edge device and its corresponding edge server, influencing the degree of time-varying data heterogeneity. A larger value of $\rho$ indicates greater variation in both data volume and data distribution throughout the learning process. Moreover, we use Adam~\cite{kingma2014adam} as the optimizer with the learning rate of 1e-3 for all trainable models in the framework. The number of communication rounds $K$ is set to 200 and all trainable parameters are initialized through Xavier~\cite{glorot2010understanding}.


\subsubsection{Baselines and Configurations}
We compare our framework with four state-of-the-art decentralized baselines, each of which can be deployed within the EC system. 
\begin{itemize}
    \item \textbf{RND} randomly selects inter-edge communication links to construct the topology in each communication round and employs a simple averaging strategy for model aggregation. The proportion of selected links to the total number of possible links is set to 0.4.
    \item \textbf{D-PSGD}~\cite{DPSGD} adopts a static ring communication topology and performs model aggregation using weighted averaging.
    \item \textbf{SGP}~\cite{pushsum} employs a time-varying directed exponential graph to represent communication connectivity and utilizes a stochastic gradient push algorithm for model aggregation. Each participant is configured with two neighbors in the communication graph.
    \item \textbf{DAC}~\cite{DAC} introduces a similarity-based sampling strategy that allows each participant to adaptively select neighbors for collaborative learning and employs averaging for model aggregation, with the number of neighbors is set to 2.
\end{itemize}

\subsubsection{Performance Metrics}
\begin{itemize}
    \item \textbf{Average Test Accuracy} (Avg. Acc) measures the mean test accuracy across all models on the edge servers.
    \item \textbf{Variance of Test Accuracy} (Var. Acc) captures the variability in test accuracy among all edge server models, indicating the consistency of model performance.
    \item \textbf{Best Test Accuracy} (B. Acc) reports the highest test accuracy achieved by any model across the edge servers.
    \item \textbf{Worst Test Accuracy} (W. Acc) records the lowest test accuracy among all models on the edge servers.
    \item \textbf{Total Energy Cost} (Tot. Cost) quantifies the cumulative energy consumption incurred during the entire learning process, measured in megajoules (MJ).
    \item \textbf{Model Transmission Energy Cost} (M.T. Cost) represents the total energy cost attributed to model transmissions throughout the learning process, measured in megajoules (MJ)
\end{itemize}

\subsection{Overall Performance (RQ1)}
\begin{table*}[]
\vspace{-2mm}
\caption{Simulation results on Fashion-MNIST and CIFAR-10 for performance evaluation.}
\resizebox{\linewidth}{!}{
\begin{tabular}{lccllllll|llllll}
\hline
\multicolumn{1}{c}{\multirow{2}{*}{DDS}} & \multirow{2}{*}{SPC}          & \multirow{2}{*}{Method} & \multicolumn{6}{c|}{Fashion-MNIST}                                                                                                                                                     & \multicolumn{6}{c}{CIFAR-10}                                                                                                                                                          \\ \cline{4-15} 
\multicolumn{1}{c}{}                     &                               &                         & \multicolumn{1}{c}{Avg. Acc} & \multicolumn{1}{c}{Var. Acc} & \multicolumn{1}{c}{B. Acc} & \multicolumn{1}{c}{W. Acc} & \multicolumn{1}{c}{Tot. Cost} & \multicolumn{1}{c|}{M.T. Cost} & \multicolumn{1}{c}{Avg. Acc} & \multicolumn{1}{c}{Var. Acc} & \multicolumn{1}{c}{B. Acc} & \multicolumn{1}{c}{W. Acc} & \multicolumn{1}{c}{Tot. Cost} & \multicolumn{1}{c}{M.T. Cost} \\ \hline
\multirow{10}{*}{$\lambda$ = 0.3}        & \multirow{5}{*}{$\rho$ = 0.5} & RND                  & 79.2$\pm$0.9                 & 1.4$\pm$0.5                  & 80.5$\pm$0.9               & 77.8$\pm$0.6               & 4.0$\pm$0.5                   & 3.3$\pm$0.5                    & 37.8$\pm$1.7                 & 1.2$\pm$0.4                  & 39.3$\pm$2.0               & 36.7$\pm$1.2               & 11.9$\pm$0.9                  & 9.4$\pm$1.1                   \\
                                         &                               & D-PSGD                  & 81.8$\pm$0.8                 & 1.1$\pm$0.4                  & 82.9$\pm$0.9               & 80.5$\pm$1.5               & 3.9$\pm$0.4                   & 3.2$\pm$0.5                    & 38.6$\pm$1.8                 & 1.7$\pm$1.0                  & 40.8$\pm$2.4               & 36.5$\pm$1.7               & 12.2$\pm$0.7                  & 9.7$\pm$0.8                   \\
                                         &                               & DAC                     & 81.1$\pm$0.7                 & 0.6$\pm$0.3                  & 81.7$\pm$0.7               & 80.4$\pm$1.0               & 4.2$\pm$0.3                   & 3.5$\pm$0.4                    & 38.1$\pm$1.5                 & 1.1$\pm$0.5                  & 39.8$\pm$1.0               & 37.7$\pm$1.8               & 11.4$\pm$0.7                  & 8.9$\pm$0.7                   \\
                                         &                               & SGP                     & 81.9$\pm$0.6                 & 0.8$\pm$0.4                  & 82.9$\pm$0.8               & 81.3$\pm$0.7               & 3.8$\pm$0.4                   & 3.1$\pm$0.3                    & 39.2$\pm$1.3                 & 1.1$\pm$0.6                  & 40.9$\pm$1.2               & 38.2$\pm$1.4               & 11.6$\pm$0.6                  & 9.1$\pm$0.7                   \\ \cline{3-15} 
                                         &                               & Ours                    & \textbf{83.9$\pm$0.5}        & \textbf{0.6$\pm$0.2}         & \textbf{84.5$\pm$0.4}      & \textbf{83.2$\pm$0.5}      & \textbf{2.3$\pm$0.3}          & \textbf{1.6$\pm$0.2}           & \textbf{41.9$\pm$1.3}        & \textbf{0.8$\pm$0.2}         & \textbf{43.0$\pm$1.3}      & \textbf{41.5$\pm$1.2}      & \textbf{7.4$\pm$0.2}          & \textbf{4.9$\pm$0.3}          \\ \cline{2-15} 
                                         & \multirow{5}{*}{$\rho$ = 0.9} & RND                  & 81.6$\pm$0.8                 & 1.0$\pm$0.4                  & 82.6$\pm$0.7               & 80.5$\pm$0.8               & 4.3$\pm$0.4                   & 3.4$\pm$0.3                    & 42.1$\pm$1.6                 & 1.0$\pm$0.4                  & 43.1$\pm$1.7               & 41.0$\pm$1.4               & 13.4$\pm$1.3                  & 9.2$\pm$1.5                   \\
                                         &                               & D-PSGD                  & 83.3$\pm$0.6                 & 1.1$\pm$0.2                  & 84.5$\pm$0.7               & 82.1$\pm$0.5               & 3.8$\pm$0.3                   & 2.9$\pm$0.4                    & 42.5$\pm$1.2                 & 1.2$\pm$0.6                  & 44.2$\pm$0.8               & 41.0$\pm$0.9               & 12.9$\pm$1.0                  & 9.7$\pm$0.8                   \\
                                         &                               & DAC                     & 83.6$\pm$0.4                 & 0.6$\pm$0.3                  & 84.4$\pm$0.3               & 83.1$\pm$0.7               & 3.9$\pm$0.5                   & 3.0$\pm$0.5                    & 43.6$\pm$0.9                 & 0.8$\pm$0.3                  & 44.2$\pm$1.0               & 42.8$\pm$0.9               & 13.0$\pm$0.9                  & 9.8$\pm$0.9                   \\
                                         &                               & SGP                     & 83.8$\pm$0.5                 & 0.7$\pm$0.2                  & 84.6$\pm$0.7               & 83.2$\pm$0.3               & 3.8$\pm$0.2                   & 2.9$\pm$0.3                    & 43.8$\pm$0.6                 & 0.9$\pm$0.2                  & 44.5$\pm$0.6               & 42.5$\pm$0.4               & 12.6$\pm$1.0                  & 9.4$\pm$1.2                   \\ \cline{3-15} 
                                         &                               & Ours                    & \textbf{85.2$\pm$0.2}        & \textbf{0.4$\pm$0.2}         & \textbf{85.7$\pm$0.3}      & \textbf{84.6$\pm$0.2}      & \textbf{2.6$\pm$0.3}          & \textbf{1.7$\pm$0.2}                    & \textbf{44.6$\pm$1.1}        & \textbf{0.7$\pm$0.2}         & \textbf{45.5$\pm$0.9}      & \textbf{43.9$\pm$1.1}      & \textbf{8.2$\pm$0.3}                   & \textbf{5.0$\pm$0.4}                   \\ \hline
\multirow{10}{*}{$\lambda$ = 0.7}        & \multirow{5}{*}{$\rho$ = 0.5} & RND                  & 83.8$\pm$0.7                 & 0.7$\pm$0.4                  & 84.4$\pm$0.8               & 83.0$\pm$0.7               & 3.9$\pm$0.5                   & 3.3$\pm$0.7                    & 45.2$\pm$1.1                 & 1.1$\pm$0.5                  & 46.2$\pm$1.2               & 43.1$\pm$1.3               & 12.8$\pm$1.2                  & 10.1$\pm$0.9                  \\
                                         &                               & D-PSGD                  & 86.0$\pm$0.4                 & 1.3$\pm$0.1                  & 87.3$\pm$0.4               & 84.3$\pm$0.8               & 4.0$\pm$0.4                   & 3.4$\pm$0.3                    & 45.6$\pm$1.0                 & 1.2$\pm$0.8                  & 46.8$\pm$1.0               & 43.7$\pm$1.2               & 12.6$\pm$0.9                  & 9.9$\pm$1.2                   \\
                                         &                               & DAC                     & 85.7$\pm$0.4                 & 0.9$\pm$0.4                  & 86.7$\pm$0.3               & 84.3$\pm$0.8               & 3.5$\pm$0.5                   & 2.9$\pm$0.4                    & 46.0$\pm$0.8                 & 1.0$\pm$0.3                  & 46.9$\pm$0.8               & 44.8$\pm$0.7               & 12.3$\pm$0.9                  & 9.6$\pm$0.8                   \\
                                         &                               & SGP                     & 86.0$\pm$0.2                 & 0.9$\pm$0.4                  & 86.7$\pm$0.3               & 84.8$\pm$0.7               & 3.7$\pm$0.3                   & 3.1$\pm$0.4                    & 46.5$\pm$0.6                 & \textbf{0.7$\pm$0.2}         & 47.3$\pm$0.7               & 45.8$\pm$0.8               & 11.0$\pm$0.6                  & 9.3$\pm$0.9                   \\ \cline{3-15} 
                                         &                               & Ours                    & \textbf{87.0$\pm$0.3}        & \textbf{0.5$\pm$0.2}         & \textbf{87.6$\pm$0.5}      & \textbf{86.4$\pm$0.4}      & \textbf{2.4$\pm$0.4}          & \textbf{1.8$\pm$0.2}           & \textbf{47.1$\pm$0.8}        & 0.8$\pm$0.2                  & \textbf{48.1$\pm$0.4}      & \textbf{46.5$\pm$0.7}      & \textbf{7.8$\pm$0.5}          & \textbf{5.1$\pm$0.3}          \\ \cline{2-15} 
                                         & \multirow{5}{*}{$\rho$ = 0.9} & RND                  & 87.2$\pm$0.5                 & 0.3$\pm$0.3                  & 87.5$\pm$0.4               & 86.7$\pm$0.8               & 4.1$\pm$0.2                   & 3.2$\pm$0.3                    & 49.3$\pm$0.9                 & 1.1$\pm$0.5                  & 50.7$\pm$1.0               & 48.1$\pm$0.9               & 12.1$\pm$0.9                  & 9.8$\pm$0.7                   \\
                                         &                               & D-PSGD                  & 87.4$\pm$0.4                 & 0.4$\pm$0.2                  & 87.8$\pm$0.4               & 87.1$\pm$0.5               & 4.2$\pm$0.5                   & 3.3$\pm$0.5                    & 49.7$\pm$0.6                 & 0.7$\pm$0.3                  & 50.5$\pm$0.8               & 48.8$\pm$0.7               & 10.2$\pm$0.7                  & 7.9$\pm$0.8                   \\
                                         &                               & DAC                     & 87.4$\pm$0.2                 & 0.5$\pm$0.1                  & 87.8$\pm$0.1               & 86.9$\pm$0.5               & 4.1$\pm$0.3                   & 3.2$\pm$0.5                    & 49.5$\pm$0.6                 & 0.8$\pm$0.3                  & 50.3$\pm$0.7               & 48.4$\pm$0.6               & 11.8$\pm$0.7                  & 9.5$\pm$0.9                   \\
                                         &                               & SGP                     & 87.6$\pm$0.5                 & \textbf{0.2$\pm$0.1}         & 87.9$\pm$0.4               & 87.4$\pm$0.7               & 4.1$\pm$0.4                   & 3.2$\pm$0.2                    & 50.1$\pm$0.5                 & 0.7$\pm$0.4                  & 51.0$\pm$0.6               & 49.4$\pm$0.6               & 11.7$\pm$0.8                  & 9.4$\pm$0.9                   \\ \cline{3-15} 
                                         &                               & Ours                    & \textbf{88.1$\pm$0.1}        & 0.3$\pm$0.2                  & \textbf{88.4$\pm$0.3}      & \textbf{87.7$\pm$0.2}      & \textbf{2.7$\pm$0.2}          & \textbf{1.8$\pm$0.2}           & \textbf{51.1$\pm$0.5}        & \textbf{0.6$\pm$0.3}         & \textbf{51.7$\pm$0.4}      & \textbf{50.3$\pm$0.7}      & \textbf{7.2$\pm$0.3}          & \textbf{4.9$\pm$0.4}          \\ \hline
\end{tabular}
}
\vspace{-5mm}
\label{tab:overview}
\end{table*}
We simulate varying levels of non-IID and time-varying data heterogeneity in the EC system by setting different values of $\lambda \in \{0.3, 0.7\}$ for DDS and $\rho\in\{0.5, 0.9\}$ for SPC. We evaluate the performance of our proposed framework using six metrics, categorized into two groups: (1) model predictive performance, measured by average test accuracy, variance of test accuracy, best test accuracy, and worst test accuracy; and (2) energy efficiency, assessed through total energy cost and model transmission energy cost. All experiments are repeated five times with different random seeds, and we report the average results along with the standard deviation. The overall performance of our framework, compared with four baselines (RND, D-PSGD, SGP, and DAC) on two datasets, is presented in Table~\ref{tab:overview}. The highest average test accuracy, best test accuracy, and worst test accuracy, as well as the lowest variance of test accuracy, total energy cost, and model transmission energy cost in each setting, are highlighted in bold.

Our proposed framework, \pname, consistently outperforms all baselines across both datasets. Specifically, it achieves improvements in average test accuracy, best test accuracy, and worst test accuracy ranging from 0.5\% to 4.7\%, 0.5\% to 4.0\%, and 0.3\% to 5.4\%, respectively. In terms of energy efficiency, \pname reduces total energy cost and model transmission energy cost by 29.0\%–45.2\% and 37.9\%–54.2\%, respectively. On average, it achieves gains of 1.9\%, 1.6\%, and 2.3\% in average, best, and worst test accuracy, while decreasing total energy cost and model transmission energy cost by 36.7\% and 46.3\%, respectively. Compared to Rand, D-PSGD, DAC and SGP, \pname achieves average improvements of 2.9\%, 1.8\%, 1.7\%, and 1.3\% in Average Test Accuracy, and reduces the Total Energy Cost by 38.8\%, 36,4\%, 36.5\%, and 34.8\%, respectively. 
In general, we observe that our framework outperforms all baselines across varying degrees of DDS and SPC, where smaller values of $\lambda$ and $\rho$ indicate more severe non-IID and time-varying data heterogeneity in the EC system. Our framework achieves a higher average improvement of 3.2\% in average test accuracy when $\lambda = 0.3$ and $\rho = 0.5$, compared to average improvements of 1.1\% to 1.9\% under other combinations of DDS and SPC settings. This demonstrates that the proposed framework enables effective collaborative learning in edge computing environments exhibiting greater data heterogeneity.

\subsection{Impact of Communication Topology Sparsity $\gamma$ (RQ2)}
\begin{figure*}[t!]
\vspace{-4mm}
    \centering
    
  \scalebox{1.0}{
  \subfloat[$\lambda = 0.3$, $\rho = 0.5$\label{p2-acm}]{%
        \includegraphics[width=0.25\linewidth]{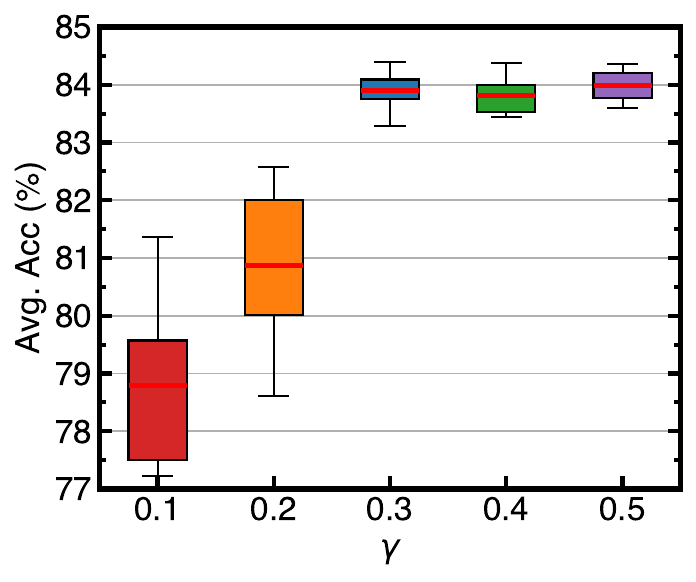}}
    \hfill
  \subfloat[$\lambda = 0.3$, $\rho = 0.9$\label{p2-dblp}]{%
        \includegraphics[width=0.25\linewidth]{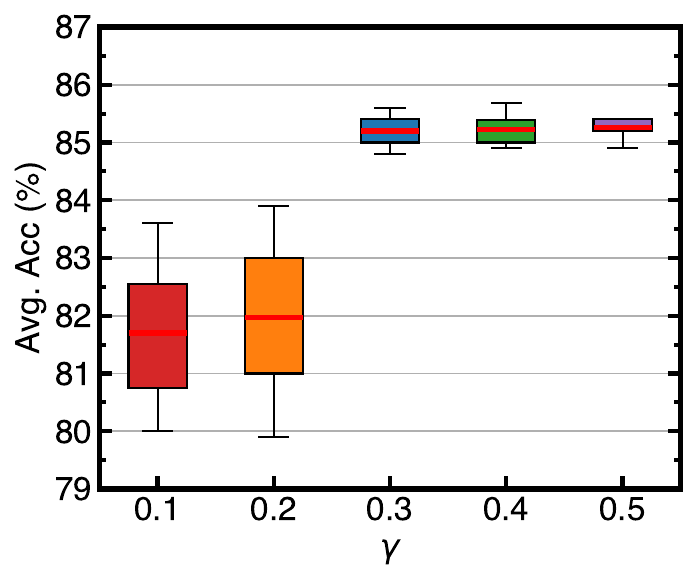}}
    \hfill
   \subfloat[$\lambda = 0.7$, $\rho = 0.5$\label{p2-imdb}]{%
       \includegraphics[width=0.25\linewidth]{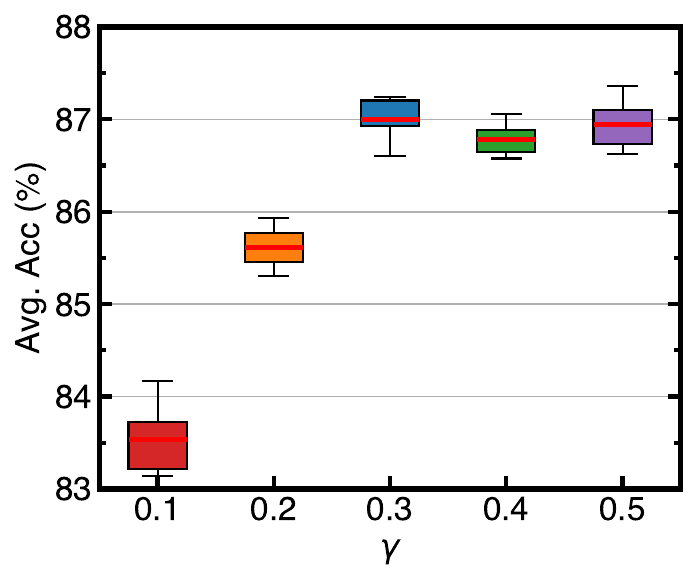}}
           \hfill
    \subfloat[$\lambda = 0.7$, $\rho = 0.9$\label{p2-imdb}]{%
       \includegraphics[width=0.25\linewidth]{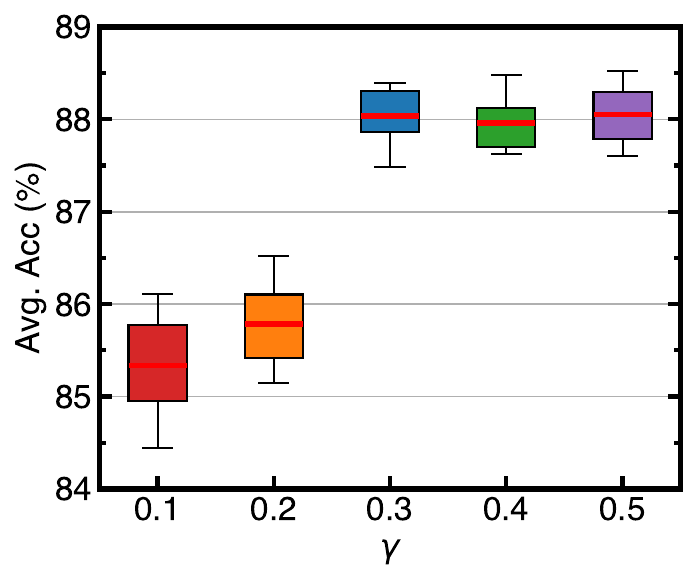}}

    }
    
  \caption{Average test accuracy of \pname under different values of $\gamma$ across various experimental settings.}
  \label{fig:sparsity} 
  \vspace{-5mm}
\end{figure*}

To answer RQ2, we conduct extensive experiments to investigate the effect of communication topology sparsity $\gamma$ in \pname under various combinations of DDS and SPC settings. \figurename~\ref{fig:sparsity} and Table~\ref{tab:sparsity} report the average test accuracy and total energy cost, respectively, as $\gamma$ ranges from 0.1 to 0.5. When $\gamma$ is small, fewer inter-edge communication links are selected to construct the topology. Although this reduces model transmission energy cost and contributes to lower total energy cost, it also limits the opportunity for edge servers to learn from its neighbors during the model aggregation process, thereby leading to degraded model predictive performance. As observed in \figurename~\ref{fig:sparsity}, increasing the inter-edge topology sparsity parameter $\gamma$ generally leads to an improvement in the average test accuracy of our framework. However, it is observed that the model's predictive accuracy improves only marginally beyond $\gamma = 0.3$. Further analysis of Table~\ref{tab:sparsity} demonstrates that the total energy cost continuously increases as $\gamma$ grows. When $\gamma = 0.3$, it is an optimal point to trade off the model performance and total energy cost in \pname. This trend is observed across experimental settings with varying levels of non-IID ($\lambda$) and time-varying ($\rho$) data heterogeneity.

\begin{threeparttable}
\caption{Total energy cost of \pname under different values of $\gamma$ across various experimental settings}
\begin{tabular}{l|cccc}
\hline
\multirow{2}{*}{\textbf{Model}} & \multicolumn{4}{c}{\textbf{Experimental Settings}} \\ \cline{2-5} 
                                & (a)        & (b)           & (c)           & (d)          \\ \hline
Our$|_{r=0.1}$                             &1.3$\pm$0.1             &1.5$\pm$0.1            &1.3$\pm$0.2            &1.5$\pm$0.1            \\
Our$|_{r=0.2}$                             &1.7$\pm$0.1             &2.1$\pm$0.2            &1.8$\pm$0.1            &2.1$\pm$0.1            \\
Our$|_{r=0.4}$                             &2.9$\pm$0.2             &3.2$\pm$0.3            &3.1$\pm$0.2            &3.4$\pm$0.1           \\
Our$|_{r=0.5}$                             &3.4$\pm$0.4             &3.7$\pm$0.4            &3.5$\pm$0.3            &4.0$\pm$0.4            \\ \hline
\textbf{Our}$|_{\boldsymbol{r=0.3}}$                    &2.3$\pm$0.3             &2.6$\pm$0.3            &2.4$\pm$0.4            &2.7$\pm$0.2            \\ \hline
\end{tabular}
\label{tab:sparsity}
    \begin{tablenotes}
	\small {
	\item{ $\text{(a)}$: $\lambda = 0.3$, $\rho = 0.5$; $\text{(b)}$: $\lambda = 0.3$, $\rho = 0.9$ }
	\item $\text{(c)}$: $\lambda = 0.7$, $\rho = 0.5$; $\text{(d)}$: $\lambda = 0.7$, $\rho = 0.9$}
    \end{tablenotes}
\vspace{-2mm}
\end{threeparttable}

\subsection{Scalability Analysis (RQ3)}
We investigate the scalability of \pname, focusing on the predictive performance of models on edge servers as the number of edge servers increases. As shown in Table~\ref{tab:scala},
we set the number of edge servers to be $\{5, 10, 20, 50\}$
in our framework and report performance metrics including  average test accuracy, variance of test accuracy, best test accuracy and worst test accuracy. As the number of edge servers increases across both datasets, the framework’s performance exhibits a slight decline, with the reduction in average test accuracy remaining within acceptable bounds, around 1.3\% in Fashion-MNIST and 1.1\% in CIFAR-10. Notably, the variance of test accuracy increases consistently as the number of edge servers grows. Such experimental results are attributed to severe non-IID skewness caused by the discrepancy in the data distributions among edge servers. In other words, the
growing number of edge servers amplifies the potential for difference and randomness in training datasets among participants, resulting in greater variance in the final predictive performance of models on the edge servers.
\begin{table}[h]
\caption{Performance comparison of different number of edge servers}
\resizebox{1.02\linewidth}{!}{
\begin{tabular}{|cc|cccc|}
\hline
\multicolumn{2}{|l|}{\multirow{2}{*}{}}                     & \multicolumn{4}{c|}{Number of Edge Servers}                                                \\ \cline{3-6} 
\multicolumn{2}{|l|}{}                                      & \multicolumn{1}{c|}{5} & \multicolumn{1}{c|}{10} & \multicolumn{1}{c|}{20} & 50 \\ \hline
\multicolumn{1}{|l|}{\multirow{4}{*}{Fashion-MNIST}}   & Avg. Acc  & \multicolumn{1}{l|}{88.1$\pm$0.1}  & \multicolumn{1}{l|}{87.6$\pm$0.8}   & \multicolumn{1}{l|}{87.2$\pm$0.7}   &86.8$\pm$0.8    \\ \cline{2-6} 
\multicolumn{1}{|l|}{}                          & Var. Acc  & \multicolumn{1}{l|}{0.3$\pm$0.2}  & \multicolumn{1}{l|}{0.7$\pm$0.2}   & \multicolumn{1}{l|}{0.9$\pm$0.3}   &1.2$\pm$0.5    \\ \cline{2-6} 
\multicolumn{1}{|l|}{}                          & B. Acc  & \multicolumn{1}{l|}{88.4$\pm$0.3}  & \multicolumn{1}{l|}{87.9$\pm$0.5}   & \multicolumn{1}{l|}{87.7$\pm$0.9}   &87.6$\pm$0.9    \\ \cline{2-6} 
\multicolumn{1}{|l|}{}                          & W. Acc & \multicolumn{1}{l|}{87.7$\pm$0.2}  & \multicolumn{1}{l|}{86.9$\pm$0.9}   & \multicolumn{1}{l|}{86.2$\pm$0.8}   &86.0$\pm$1.2    \\ \hline\hline
\multicolumn{1}{|l|}{\multirow{4}{*}{CIFAR-10}} & Avg. Acc  & \multicolumn{1}{l|}{51.1$\pm$0.5}  & \multicolumn{1}{l|}{50.7$\pm$0.9}   & \multicolumn{1}{l|}{50.4$\pm$0.8}   &50.0$\pm$0.9    \\ \cline{2-6} 
\multicolumn{1}{|l|}{}                          & Var. Acc  & \multicolumn{1}{l|}{0.6$\pm$0.3}  & \multicolumn{1}{l|}{0.9$\pm$0.4}   & \multicolumn{1}{l|}{1.0$\pm$0.3}   & 1.0$\pm$0.4   \\ \cline{2-6} 
\multicolumn{1}{|l|}{}                          & B. Acc  & \multicolumn{1}{l|}{51.7$\pm$0.4}  & \multicolumn{1}{l|}{51.3$\pm$0.4}   & \multicolumn{1}{l|}{50.9$\pm$0.6}   &50.7$\pm$0.8    \\ \cline{2-6} 
\multicolumn{1}{|l|}{}                          & W. Acc & \multicolumn{1}{l|}{50.3$\pm$0.7}  & \multicolumn{1}{l|}{49.4$\pm$0.6}   & \multicolumn{1}{l|}{49.1$\pm$0.8}   &48.8$\pm$0.9    \\ \hline
\end{tabular}
}
\label{tab:scala}
\vspace{-5mm}
\end{table}

\subsection{Sensitivity Analysis (RQ4)}
\begin{figure}[t]
\vspace{-2mm}
    \centering
    
    \includegraphics[width=1\linewidth]{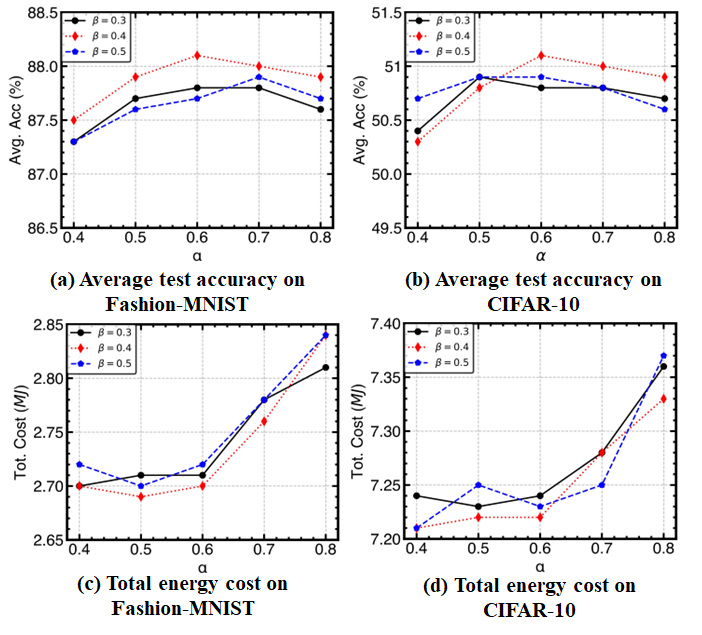}
  \caption{Performance of \pname as a function of $\alpha$ for several values of $\beta$ for two datasets.}
  \label{fig:sens_a}
  \vspace{-3mm}
\end{figure}

\begin{figure}[t]
\vspace{-2mm}
    \centering
    
    \includegraphics[width=1\linewidth]{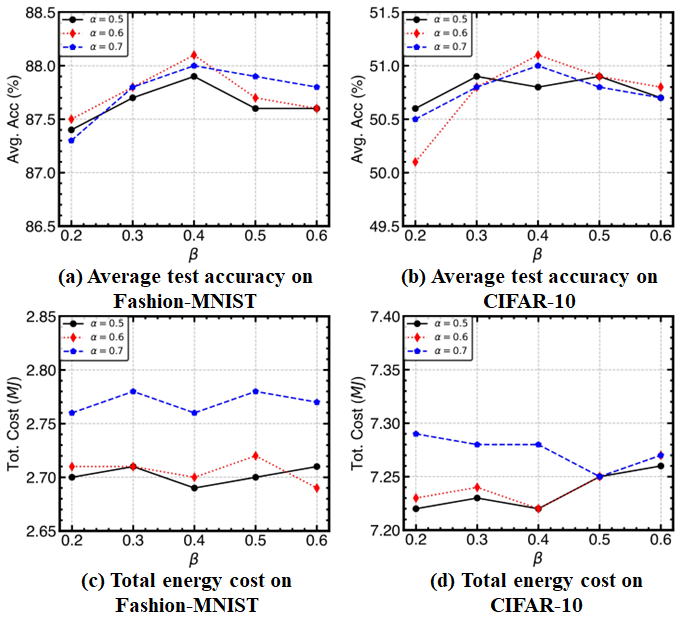}
    \vspace{-4mm}

  \caption{Performance of \pname as a function of $\beta$ for several values of $\alpha$ for two datasets.}
  \label{fig:sens_b} 
  \vspace{-3mm}
\end{figure}

In this section, we study the impact of key hyperparameters in \pname, namely $\alpha$ and $\beta$, as illustrated in \figurename~\ref{fig:sens_a} and \figurename~\ref{fig:sens_b}. Specifically, $\alpha$ is a weight hyperparameter that controls the tradeoff between the normalized energy cost factor $S^{c}_k$ and the model performance factor $S^{c}_k$ in the utility calculation (See Eq.~\ref{eq:ture utility}). The hyperparameter $\beta$, on the other hand, balances the effects of model informativeness and training data size of models during the importance-aware model aggregation (See Eq.\ref{eq:agg_w_assign}). \figurename~\ref{fig:sens_a} and~\ref{fig:sens_b} illustrate the overall performance of \pname influenced by $\alpha$ and $\beta$. 

\figurename~\ref{fig:sens_a} illustrates the influence of $\alpha$ on average test accuracy and total energy cost across two datasets. $\beta$ is fixed at 0.3, 0.4, and 0.5, while $\alpha$ varies from 0.4 to 0.8. When $\alpha$ is small, the normalized model performance factor $S^{p}_k$ becomes trivial in the utility calculation of communication links, leading the topology optimization to focus on minimizing total energy cost throughout the learning process, which consequently degrades the model’s predictive performance. As $\alpha$ increases to around 0.6, $S^{p}_k$ accounts for the appropriate proportion of the utility, resulting in the highest average test accuracy under this configuration while maintaining a relatively low energy cost. When $\alpha$ continually increases to 0.8, average test accuracy drops slightly, but the total energy cost rises sharply. This clearly demonstrates that the dominance of $S^{p}_k$ compromises the energy efficiency of \pname.

On the other hand, \figurename~\ref{fig:sens_b} describes the influence of $\beta$. When $\beta$ is very small, the effect of training data size dominates, resulting in models trained on larger datasets being assigned greater aggregation weights and thus contributing more during the aggregation process. However, neglecting model importance of models compromises predictive performance, as illustrated in \figurename~\ref{fig:sens_b} (a) and (b). When $\beta$ continually increases to 0.6, the dominance of model importance does not consistently enhance predictive performance, as the time-varying data heterogeneity introduces variance during local training. Consequently, ignoring the effect of training data size may compromise the effectiveness of model aggregation. With $\beta$ ranging in $[0.2,0.6]$, the average test accuracy reaches the highest when $\beta = 0.4$. 

\begin{table}[h]
\caption{Performance of \pname for ablation study.}
\resizebox{\linewidth}{!}{
\begin{tabular}{|l|llllll|}
\hline
\multirow{2}{*}{\textbf{Model}} & \multicolumn{6}{c|}{\textbf{Fashion-MNIST}}                                                                                                                                                                                                                \\ \cline{2-7} 
                       & \multicolumn{1}{c|}{\textbf{Avg. Acc}} & \multicolumn{1}{c|}{\textbf{Var. Acc}} & \multicolumn{1}{c|}{\textbf{B. Acc}} & \multicolumn{1}{c|}{\textbf{W. Acc}} & \multicolumn{1}{c|}{\textbf{Tot. Cost}} & \multicolumn{1}{c|}{\textbf{M.T. Cost}} \\ \hline
\textbf{Ours}          & \multicolumn{1}{l|}{83.9$\pm$0.5}                  & \multicolumn{1}{l|}{0.6$\pm$0.2}                  & \multicolumn{1}{l|}{84.5$\pm$0.4}                & \multicolumn{1}{l|}{83.2$\pm$0.5}                & \multicolumn{1}{l|}{2.3$\pm$0.3}                   &1.6$\pm$0.2                                         \\ \hline
\textbf{-w/o UTC}     & \multicolumn{1}{l|}{81.2$\pm$0.4}                  & \multicolumn{1}{l|}{0.8$\pm$0.3}                  & \multicolumn{1}{l|}{82.1$\pm$0.8}                & \multicolumn{1}{l|}{80.2$\pm$0.4}                & \multicolumn{1}{l|}{2.8$\pm$0.3}                   &2.1$\pm$0.3                                         \\ \hline
\textbf{-w/o IMA}      & \multicolumn{1}{l|}{80.4$\pm$0.4}                  & \multicolumn{1}{l|}{1.0$\pm$0.3}                  & \multicolumn{1}{l|}{81.4$\pm$0.5}                & \multicolumn{1}{l|}{79.3$\pm$0.4}                & \multicolumn{1}{l|}{2.4$\pm$0.1}                   & 1.7$\pm$0.2                                        \\ \hline\hline
\multirow{2}{*}{\textbf{Model}} & \multicolumn{6}{c|}{\textbf{CIFAR-10}}                                                                                                                                                                                                                     \\ \cline{2-7} 
                       & \multicolumn{1}{c|}{\textbf{Avg. Acc}} & \multicolumn{1}{c|}{\textbf{Var. Acc}} & \multicolumn{1}{c|}{\textbf{B. Acc}} & \multicolumn{1}{c|}{\textbf{W. Acc}} & \multicolumn{1}{c|}{\textbf{Tot. Cost}} & \multicolumn{1}{c|}{\textbf{M.T. Cost}} \\ \hline
\textbf{Ours}          & \multicolumn{1}{l|}{41.9$\pm$0.5}                  & \multicolumn{1}{l|}{0.8$\pm$0.2}                  & \multicolumn{1}{l|}{43.0$\pm$1.3}                & \multicolumn{1}{l|}{40.8$\pm$1.2}                & \multicolumn{1}{l|}{7.4$\pm$0.2}                   &4.9$\pm$0.3                                         \\ \hline
\textbf{-w/o UTC}     & \multicolumn{1}{l|}{39.8$\pm$1.6}                  & \multicolumn{1}{l|}{0.8$\pm$0.5}                  & \multicolumn{1}{l|}{40.8$\pm$1.4}                & \multicolumn{1}{l|}{38.6$\pm$1.6}                & \multicolumn{1}{l|}{8.1$\pm$0.5}                   &5.6$\pm$0.5                                          \\ \hline
\textbf{-w/o IMA}      & \multicolumn{1}{l|}{38.9$\pm$2.2}                  & \multicolumn{1}{l|}{1.1$\pm$0.5}                  & \multicolumn{1}{l|}{40.1$\pm$1.7}                & \multicolumn{1}{l|}{37.7$\pm$2.0}                & \multicolumn{1}{l|}{7.5$\pm$0.4}                   &5.0$\pm$0.5                                         \\ \hline
\end{tabular}
}
\vspace{-5mm}
\label{tab:ablation}
\end{table}

\subsection{Ablation Study (RQ5)}
To validate the effectiveness of the key components in \pname, we design two variants and compare their performance against \pname on two datasets, as reported in Table~\ref{tab:ablation}. Specifically, we investigate the contributions of utility-based topology construction (UTC) module and the importance-aware model aggregation (IMA) to the performance of \pname. \pname w/o UTC variant constructs the communication topology using a random strategy, where the selected communication links are chosen randomly, with the total number of links matching that of \pname. The results show that \pname outperforms \pname w/o UTC by margins of 2.1\%–2.7\%, 2.2\%–2.4\%, and 2.2\%–2.9\% in terms of average, best, and worst test accuracy, respectively, across both datasets. In addition, \pname reduces the total energy cost and model transmission energy cost by an average of 13.2\% and 18.9\%, respectively, compared to \pname w/o UTC. These results are attributed to UTC’s capacity to dynamically construct inter-edge topologies by leveraging the observed utility of communication links, which quantifies the historical impact of communication link selection on both model performance improvement and energy cost, to guide their future selection probabilities. \pname w/o IMA excludes model importance evaluation, assigning equal weight to each model during the aggregation process. The three metrics related to test accuracy of \pname w/o IMA drop significantly on both datasets, while the variance of test accuracy increases. This indicates that the design of the importance-aware model aggregation in \pname enhances model generalization in decentralized collaborative learning within heterogeneous EC systems.


\section{Conclusion}
\label{sec:7}
In this paper, we propose a green DFL framework to enable efficient collaborative learning in heterogeneous EC systems, with the goal of improving model performance while reducing cumulative energy cost during the learning process. Within this framework, topology construction is formulated as a joint optimization problem, where each topology connection's dual impact on both model performance improvement and energy consumption is quantified by a novel evaluation metric, termed utility. To solve this problem, UTC first estimates the utility of communication links based on historical data related to model performance and various types of energy cost. The optimal communication topology is then generated based on the estimated utility, aiming to balance model performance and energy efficiency. Subsequently, DCMU is conducted among participants, where a quality-aware model aggregation strategy mitigates the degradation in model performance caused by training data heterogeneity. We conduct comprehensive experiments on two datasets and compare our framework with four state-of-the-art baselines to demonstrate its effectiveness. The results show that the proposed framework consistently outperforms existing approaches in both model performance and energy efficiency throughout the collaborative learning process. In future work, we will investigate optimal communication topology construction for deploying DFL with significantly larger trainable parameters under dynamic edge conditions and limited energy resources, aiming to further enhance the energy efficiency of the framework.

\bibliographystyle{IEEEtran}
\bibliography{IEEEabrv,ref}
\end{document}